\documentclass[]{main}

\title{\vspace{-1pt}Seirênes: Adversarial Self-Play with Evolving Distractions for LLM Reasoning}

\author[1]{Chi Zhang}
\author[2,\dagger]{Haibo Qiu}
\author[3]{Qiming Zhang}
\author[3]{Yufei Xu}
\author[4]{Xinbo Gao}
\author[1,*]{Jing Zhang}

\affiliation[1]{School of Computer Science, Wuhan University}
\affiliation[2]{Meituan Inc}
\affiliation[3]{Independent Researchers}
\affiliation[4]{Xidian University}
\contribution[\dagger]{Project Leader}
\contribution[*]{Corresponding author}

\code{https://github.com/MiliLab/Seirenes}
\correspondence{\email{jingzhang.cv@gmail.com}}
\date{\today}

\usepackage{amsmath,amsfonts,amssymb,amsthm}
\usepackage{algorithm}
\usepackage{algorithmic}
\usepackage{environ}
\usepackage{enumitem}
\usepackage{marvosym}
\usepackage{nicefrac}

\setlist[itemize]{leftmargin=*, itemsep=2pt, topsep=3pt}
\setlist[enumerate]{leftmargin=*, itemsep=2pt, topsep=3pt}

\newcommand{\headbf}[1]{\textcolor{sysugreen}{\textbf{#1}}}
\renewcommand{\paragraph}[1]{\vspace{4pt}\noindent\headbf{#1.}\hspace{0.3em}}

\let\settitleabstract\abstract
\newcommand{\storedtitleabstract}{}
\RenewEnviron{abstract}{%
  \global\let\storedtitleabstract\BODY
  \settitleabstract{\storedtitleabstract}%
}

\begin{document}

% \begin{abstract}

% We present \textbf{Seirênes} (\textbf{S}elf-\textbf{I}mproving
% \textbf{R}easoning via \textbf{E}volving \textbf{N}egative hints), a
% self-play RL framework that turns contextual interference from a failure
% mode of LLM reasoning into a training signal for self-improvement. While
% reinforcement learning with verifiable rewards has substantially
% advanced reasoning capabilities, even extensively RL-trained models
% remain fragile under misleading context, suggesting that clean benchmark
% gains may overstate robust reasoning ability. Seirênes harnesses this
% vulnerability through a shared-parameter adversarial self-play loop.
% Within this framework, a single model is trained both to construct
% plausible but misleading hints designed to derail its own reasoning,
% and to see through such interference and uncover the true underlying
% logic of each problem. By pitting these competing objectives against
% each other, Seirênes forces the
% model to move beyond superficial pattern matching and memorization,
% anchoring its capabilities in robust, intrinsic reasoning. This
% continuous interaction sustains an informative co-evolutionary
% curriculum as the model improves. Across model scales, training with Seirênes
% delivers consistent gains over strong GRPO baselines on mathematical
% reasoning and substantially reduces degradation under adversarial
% distractors. Our code will be available.

% \end{abstract}

% --- Alternative abstract draft: Seirênes + adversarial hints ---
\begin{abstract}

We present \textbf{Seirênes}, a self-play RL framework that transforms
contextual interference from a failure mode of LLM reasoning into an internal
training signal for co-evolving more resilient reasoners. While
RL with verifiable rewards has significantly advanced reasoning capabilities,
models can still exhibit fragility when encountering non-idealized contexts:
scenarios characterized by superfluous information, tangential instructions,
or incidental correlations that differ from the clean distributions typical of
standard benchmarks.
Seirênes harnesses this vulnerability through a parameter-shared and adversarial self-play loop. Within this framework, a single model is trained to both \textit{construct plausible yet distracting contexts that expose its own reasoning blind spots}, and \textit{solve problems by discerning the essential task from these perturbations to recover the core underlying logic}. By pitting these competing objectives
against each other, Seirênes compels the model to move beyond superficial
pattern matching and anchors its capabilities in robust underlying reasoning.
This continuous interaction sustains an informative co-evolutionary curriculum as the model improves. Across seven mathematical reasoning benchmarks and model scales from 4B to 30B, Seirênes achieves average gains of +10.2, +9.1, and +7.2 points. 
% As a cross-model diagnostic, 
Besides, distracting contexts produced by the 4B Seirênes model reduce the accuracy of top-tier closed-source models (GPT and Gemini) by roughly 4--5 points, revealing Seirênes' general ability to uncover reasoning models' blind spots.
\end{abstract}

\maketitle
\renewcommand{\thefootnote}{\arabic{footnote}}
\setcounter{footnote}{0}

\section{Introduction}

\begin{list}{}{\setlength{\leftmargin}{1em}\setlength{\rightmargin}{1em}}
\item\relax\itshape
``You will come to the Seirênes first of all; they bewitch any mortal who approaches them.''
\par\noindent\hfill\normalfont\upshape --- Homer, \emph{Odyssey} XII, trans. Shewring~\cite{homer2008odyssey}
\end{list}
Reinforcement learning with verifiable rewards (RLVR) has become the central driver of progress in LLM reasoning~\cite{lambert2024tulu}. Propelled by a continuous streak of breakthroughs across complex reasoning domains, models have achieved competitive results in Olympiad-level mathematics and programming contests that once required years of human expertise~\cite{guo2025deepseek,shao2024deepseekmath,jaech2024openai,hu2025open,luo2025deepcoder,bai2025intern}.

This rapid progress has renewed interest in self-play training for reasoning
models. By coupling complementary roles within a training loop, such systems can
dynamically generate curricula and supervision signals, echoing the historic
success of AlphaGo~\cite{silver2016mastering}. Recent work such as
R-Zero~\cite{huang2025r}, Absolute Zero~\cite{zhao2025absolute}, and
SPICE~\cite{liu2025spice} demonstrates the power of this paradigm. Existing
methods typically instantiate this game at the level of task generation or
solution verification: one role proposes problems, programs, or supervision,
while another learns to solve them.

We explore a complementary form of self-play that leaves the task itself fixed
but evolves the context surrounding it. This direction is motivated by prior
evidence that strong reasoners remain brittle under seemingly minor contextual
changes: structural reformulations of math problems reduce accuracy by
27--31\%~\cite{huang2025math}, biasing context can silently redirect
chain-of-thought trajectories~\cite{turpin2023language}, and irrelevant retrievals can reduce accuracy from 96\% to 65\%~\cite{Chen2023BenchmarkingLL}.
Taken together, these observations suggest that contextual interference can
serve as more than an evaluation-time stress test. By exposing reasoning shortcuts that clean tasks may leave hidden, distracting contexts allow the curriculum to shift from \emph{which} problems a model can solve to \emph{how} it reasons through a problem. If a model can learn to generate such
contexts against itself, those contexts can become an online training signal for
improving the standalone reasoner.

We propose \textbf{Seirênes}, a self-play RL framework that turns contextual
interference from a failure mode of LLM reasoning into an internal training
signal driven by adversarial co-evolution. The name evokes the Seirênes of Greek mythology, whose alluring songs drew sailors away from their course; in our setting, the lure takes the form of plausible but misleading contextual hints. A single shared-parameter policy plays both an \textbf{\emph{Adversary}}, which
generates plausible but misleading \emph{adversarial hints}, and a
\textbf{\emph{Reasoner}}, which must solve the original problem under the perturbed
context. The Adversary is rewarded for derailing the Reasoner's solution trajectory,
while the Reasoner is trained to anchor its deduction and recover the core problem logic under such interference. This adaptive training curriculum encourages Seirênes to continuously expose and solve new reasoning failures, thereby avoiding fragile reasoning shortcuts and improving robustness.
% As both roles improve, the loop continually converts exposed reasoning failures into an adaptive curriculum. 
Crucially, Seirênes modifies only the input context: the underlying task, verifier, and test-time interface remain unchanged, so the resulting model is optimized as a stronger standalone reasoner rather than as a special-purpose robustness module.

In summary, our main contributions are as follows:

\begin{itemize}
    \item We introduce \textbf{Seirênes}, a self-play RL framework in which a
  single model co-evolves as both an \emph{Adversary} that generates
  plausible but misleading adversarial hints and a \emph{Reasoner} that
  learns to recover correct reasoning paths under such interference. This
  interaction turns contextual brittleness into an adaptive training
  curriculum.

    \item We develop a parameter-shared design that makes this
  adversarial loop trainable end to end, combining role-aware rewards,
  paired clean and hint-conditioned rollouts, role-specific update streams,
  and an efficient orchestration strategy that keeps 
  cost tractable at scale.

%     \item Extensive experiments across model scales show that Seirênes 
% delivers consistent gains over both standard RL and cooperative-hint 
% baselines. Controlled analyses attribute this success to the dynamic 
% adversarial loop, ruling out extra compute or static training curricula, 
% while additional evaluations confirm transfer to broader 
% contextual perturbations.

    \item Extensive experiments show that Seirênes delivers consistent gains over both standard RL and cooperative-hint baselines on diverse mathematical reasoning benchmarks, generalizing to broader contextual perturbations. Controlled analyses attribute these gains to the dynamic adversarial loop, ruling out extra compute or static training curricula.
\end{itemize}

\section{Related work}

\noindent\textbf{RLVR for Reasoning.} RLVR is the central post-training paradigm for reasoning LLMs~\cite{lambert2024tulu,li2025system}. While early studies demonstrate the efficacy of verifiable training signals~\cite{li2025system,wang2024math}, DeepSeekMath~\cite{shao2024deepseekmath} marks a major algorithmic milestone by introducing GRPO, which is now foundational for open-weight RLVR research. OpenAI o1~\cite{jaech2024openai} and DeepSeek-R1~\cite{guo2025deepseek} catalyze the current wave of reasoning models by demonstrating the massive potential of test-time scaling and long-horizon RL. DeepSeek-R1-Zero, in particular, advances the field by demonstrating that pure RL can directly elicit strong reasoning capabilities from base models without supervised fine-tuning~\cite{guo2025deepseek}. A subsequent wave of research focuses on refining these pipelines, tackling core optimization challenges like training stability, value estimation, and reproducibility~\cite{yu2025dapo,zheng2025group,hu2025open,yue2025vapo,liu2025understanding,liu2025prorl,he2025justrl}. As baseline reasoning capabilities rapidly expand, a long-standing question gains immediate relevance: can models move beyond being optimized by static pipelines, and actively participate in generating the challenges and signals that drive their own continuous improvement?

\noindent\textbf{Self-Evolving Reasoning.} Recent literature explores this possibility by enabling models to synthesize their own curricula via autonomous self-play~\cite{huang2025r,zhao2025absolute,liu2025spice,yu2025guided,xia2025agent0}. For instance, R-Zero investigates ungrounded challenger-solver co-evolution starting from zero data~\cite{huang2025r}, Absolute Zero anchors self-play in executable environments with verifiable feedback~\cite{zhao2025absolute}, and SPICE introduces corpus-grounded self-play to sustain improvements through document-level information asymmetry~\cite{liu2025spice}.  Subsequent extensions incorporate minimal human anchors (e.g., R-Few~\cite{yu2025guided}) or tool integration (e.g., Agent0~\cite{xia2025agent0}) to advance this paradigm. Similar concepts also appear in multimodal domains~\cite{liu2025agent0, he2025visplay}. Our work distinguishes itself by shifting the focus of co-evolution from task generation to the underlying reasoning process itself, leveraging the adversarial context surrounding a task as the primary catalyst for continued improvement.

\noindent\textbf{Hint-Augmented RL and Intervention-Based Guidance.} A closely related line of work augments RL training with hints, scaffolds, interventions, or other auxiliary guidance~\cite{xia2026learning,liao2026self,zhang2025stephint,zhang2025scaf,yang2026int,yan2025learning}. HiLL and SAGE use adaptive helpful hints to densify reward or reshape the rollout distribution~\cite{xia2026learning,liao2026self}. StepHint derives multi-level stepwise hints from stronger reasoning traces~\cite{zhang2025stephint}. Neighboring methods like Scaf-GRPO, InT and LUFFY further improve training through localized interventions or guidance from stronger external traces~\cite{zhang2025scaf,yang2026int,yan2025learning}. In stark contrast to these methods, our framework is fundamentally \emph{adversarial} and \emph{co-evolutionary}. A shared-parameter model competes against itself---simultaneously playing the \emph{Adversary} to synthesize adversarial hints, and the \emph{Reasoner} to overcome them. This attacker--target structure also connects to classical adversarial 
training~\cite{madry2017towards,jia2017adversarial} and LLM 
red-teaming~\cite{perez2022red,ganguli2022red,zou2023universal}. These areas primarily target robustness evaluation, attack defense, or
safety tuning; we discuss the connection further in
Appendix~\ref{app:related-extended}.

% ============================================================
%  Preliminaries
% ============================================================

\section{Preliminaries}
\label{sec:prelim}
We build on Group Relative Policy Optimization (GRPO)~\cite{shao2024deepseekmath}, a
critic-free policy gradient algorithm that replaces the learned value baseline with
group-relative reward normalization.
Given a question $q$, the behavior policy $\pi_{\theta_{\mathrm{old}}}$ draws
$G$ rollouts:
\begin{equation}
  \{o_i\}_{i=1}^{G} \sim \pi_{\theta_{\mathrm{old}}}(\cdot \mid q),
  \qquad
  R_i = R(q, o_i) \in \{0, 1\},
  \label{eq:rollout}
\end{equation}
where $R(q, o_i)$ is a binary verifiable reward (correct / incorrect).
The advantage of each response is estimated by standardizing rewards within the group:
\begin{equation}
  \hat{A}_{i} = \frac{R_i - \frac{1}{G}\sum_{j=1}^{G} R_j}
                     {\mathrm{std}\!\left(\{R_j\}_{j=1}^{G}\right) + \epsilon}.
  \label{eq:advantage}
\end{equation}
All tokens within response $o_i$ share the same outcome-level advantage
$\hat{A}_{i,t} = \hat{A}_i$~\cite{shao2024deepseekmath,yu2025dapo}.
The policy is then updated by maximizing the clipped surrogate objective with a
KL-divergence penalty:
\begin{equation}
  J_{\mathrm{GRPO}}(\theta)
  = \mathbb{E}\!\left[
      \frac{1}{G}\sum_{i=1}^{G}
      \frac{1}{|o_i|}\sum_{t=1}^{|o_i|}
      \min\!\left(
        r_{i,t}(\theta)\,\hat{A}_i,\;
        \mathrm{clip}(r_{i,t}(\theta),\,1{-}\varepsilon,\,1{+}\varepsilon)\hat{A}_i
      \right)
      - \beta\,D_{\mathrm{KL}}(\pi_\theta \| \pi_{\mathrm{ref}})
    \right],
  \label{eq:grpo}
\end{equation}
where the per-token importance ratio is
$r_{i,t}(\theta) =
 \pi_\theta(o_{i,t} \mid q, o_{i,<t}) \,/\, \pi_{\theta_{\mathrm{old}}}(o_{i,t} \mid q, o_{i,<t})$,
$\varepsilon$ is the clipping threshold, and $\beta$ controls the strength of
the KL penalty toward the reference policy $\pi_{\mathrm{ref}}$. When applying
GRPO to a finite rollout group below, we write
$J_{\mathrm{GRPO}}(\theta;\mathcal{B})$ for the empirical version of
Eq.~\ref{eq:grpo} evaluated on batch $\mathcal{B}$.

% ============================================================
%  Method
% ============================================================

\section{Method}
\label{sec:method}

\subsection{Role-conditioned rollout construction}
\label{sec:Seirênes-framework}

\begin{figure*}[t]
  \centering
  \includegraphics[width=\linewidth]{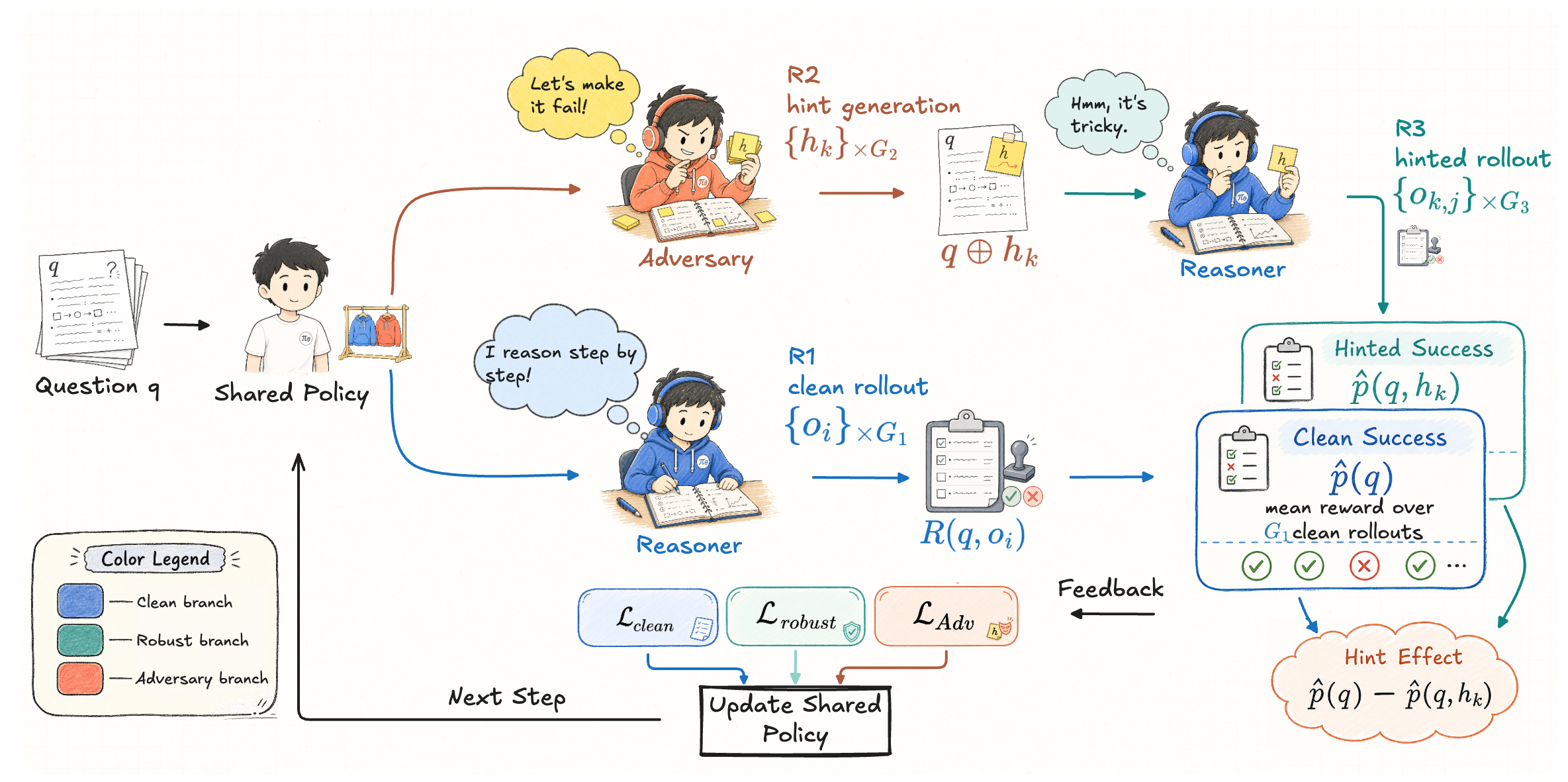}
  \caption{Seirênes: A single-policy internal arms race.}
  \vspace{-2mm}
  \label{fig:Seirênes-pipeline}
  \vspace{-4mm}
\end{figure*}

To instantiate Seirênes, a single policy with parameters $\theta$ is prompted into
two role-conditioned distributions: the Reasoner
$\pi_\theta^{\mathrm{R}}$ and the Adversary
$\pi_\theta^{\mathrm{Adv}}$.  For each training question $q$, Seirênes constructs a
paired rollout bundle that compares the Reasoner's behavior on the original
problem with and without an adversarially generated hint, as illustrated in
Fig.~\ref{fig:Seirênes-pipeline}. This process unfolds in three stages: obtaining a clean reference behavior (R1), synthesizing adversarial hints (R2), and evaluating the Reasoner under each of these interferences (R3). Together, this structured sampling yields a comprehensive trajectory bundle that isolates the causal impact of the adversarial context.

\noindent\textbf{R1: Clean rollout.} The policy first acts as the Reasoner on the original question $q$.  It samples
$G_1$ clean answer trajectories and computes their corresponding verifier rewards:
\begin{equation}
  \{o_i^{(1)}\}_{i=1}^{G_1}
  \sim
  \pi_{\theta_{\mathrm{old}}}^{\mathrm{R}}(\cdot \mid q),
  \qquad
  R_i^{(1)} = R(q,o_i^{(1)}).
  \label{eq:r1-clean}
\end{equation}
The empirical clean success rate is defined as:
\begin{equation}
  \hat p(q)
  =
  \frac{1}{G_1}\sum_{i=1}^{G_1} R(q,o_i^{(1)}).
  \label{eq:r1-clean-estimate}
\end{equation}
This establishes the Reasoner's unperturbed baseline capability.  The estimate
$\hat p(q)$ serves as an online difficulty signal for the current Reasoner: low
values indicate that the model has not yet reliably solved the clean problem,
whereas high values make subsequent failures under adversarial context more
attributable to the hint rather than to baseline task difficulty.

\noindent\textbf{R2: Adversarial hint generation.} Operating as the Adversary, the shared policy samples $G_2$ hint candidates for the same question:
\begin{equation}
  \{h_k\}_{k=1}^{G_2}
  \sim
  \pi_{\theta_{\mathrm{old}}}^{\mathrm{Adv}}(\cdot \mid q).
  \label{eq:r2-hints}
\end{equation}
The Adversary is specifically prompted to generate context that is natural and locally plausible, deliberately steering the Reasoner toward an incorrect analytical path. Details of the exact role prompt are deferred to Appendix~\ref{app:prompt}.

\noindent\textbf{R3: Hint-conditioned rollout.} For each adversarial hint $h_k$, we construct an augmented prompt $q\oplus h_k$.  The Reasoner role then answers this
context, sampling $G_3$ trajectories per hint, and we obtain their rewards using the same ground-truth answer, as the hint introduces interference without altering the underlying problem:
\begin{equation}
  \{o_{k,j}^{(3)}\}_{j=1}^{G_3}
  \sim
  \pi_{\theta_{\mathrm{old}}}^{\mathrm{R}}(\cdot \mid q\oplus h_k),
  \qquad
  R_{k,j}^{(3)} = R(q,o_{k,j}^{(3)}),
  \label{eq:r3-hinted}
\end{equation}
To quantify the Reasoner's robustness against this specific interference, we aggregate these rewards into an empirical hint-conditioned success rate:
\begin{equation}
  \hat p(q,h_k)
  =
  \frac{1}{G_3}\sum_{j=1}^{G_3} R(q,o_{k,j}^{(3)}).
  \label{eq:r3-hinted-estimate}
\end{equation}
Together with the clean estimate in Eq.~\ref{eq:r1-clean-estimate}, this
quantity establishes the exact performance drop needed to compute the adversarial reward.

\subsection{Branch-specific training signals and update rules}
\label{sec:branch-signals}
The rollout bundle provides three trajectory types with distinct credit semantics: clean Reasoner answers, Adversary hints, and hinted Reasoner answers. We now define the branch-specific training signals and update rules associated with each stream.

\noindent\textbf{Clean reasoning branch.} The clean branch applies GRPO to the R1 answer trajectories with verifier
rewards $R_i^{(1)}$:
\begin{equation}
  \mathcal{L}_{\mathrm{clean}}
  =
  -J_{\mathrm{GRPO}}
  \!\left(\theta;\{(q,o_i^{(1)},R_i^{(1)})\}_{i=1}^{G_1}\right).
  \label{eq:clean-loss}
\end{equation}
By preserving ordinary RLVR training on the original task, Seirênes keeps the Reasoner anchored to its core problem-solving capabilities. This ensures the adversarial-context branches provide auxiliary robustness pressure without distorting the underlying task distribution.

\noindent\textbf{Adversary branch.} The Adversary is trained on the hint tokens generated in R2.  We measure hint
effectiveness by the performance drop from clean to hint-conditioned success.  At the
population level, for the behavior policy that generates the paired bundle, we define
\begin{equation}
  p(q)
  =
  \mathbb{E}_{o\sim\pi_{\theta_{\mathrm{old}}}^{\mathrm{R}}(\cdot\mid q)}
  \!\left[R(q,o)\right],
  \qquad
  p(q,h_k)
  =
  \mathbb{E}_{o\sim\pi_{\theta_{\mathrm{old}}}^{\mathrm{R}}(\cdot\mid q\oplus h_k)}
  \!\left[R(q,o)\right].
  \label{eq:clean-hinted-success}
\end{equation}
Their difference defines the hint-effectiveness gap: 
$\Delta(q,h_k)=p(q)-p(q,h_k).$ In practice, Seirênes estimates this quantity empirically from the paired rollout
bundle:
\begin{equation}
  \hat R_{\mathrm{Adv}}(q,h_k)
  =
  \hat\Delta(q,h_k)
  =
  \hat p(q)-\hat p(q,h_k).
  \label{eq:adv-reward}
\end{equation}
Positive values indicate that the hint degrades the current Reasoner relative to its
clean behavior.  Beyond identifying degradation, this
empirical gap provides a naturally bounded and asymmetric credit signal.  Since $\hat p(q,h_k)\in[0,1]$, the reward lies in
$[-(1-\hat p(q)),\hat p(q)]$. This structural property elegantly prevents common failure modes in adversarial training due to the inherent design of the reward formulation, without resorting to hand-tuned curricula or auxiliary stabilization tricks:
\begin{itemize}
    \item \textbf{The ``Too Hard'' Regime (Low $\hat p(q)$):} If the model already struggles with the clean problem, a hinted failure provides weak evidence that the hint caused the disruption. The available positive credit is naturally throttled.
    \item \textbf{The ``Too Easy'' Regime (High $\hat p(q)$, High $\hat p(q,h_k)$):} If the Reasoner easily solves the clean problem and resists the hint, both rates are high, attenuating the update magnitude. This prevents a one-sided dominance collapse---a common self-play instability where the Adversary is over-penalized on questions that are too easy for the Reasoner.
\end{itemize}

Consequently, strong positive credit is exclusively concentrated on the intended sweet spot: questions that the Reasoner can solve cleanly but fails on when misled.

To optimize the Adversary, the number of hint candidates $G_2$ controls a trade-off between rollout cost and the resolution of within-question ranking. In cost-efficient settings---which we adopt as the default configuration for our main experiments---Seirênes applies a REINFORCE-style token update:
\begin{equation}
  \mathcal{L}_{\mathrm{Adv}}
  =
	  -
	  \frac{1}{G_2}\sum_{k=1}^{G_2}
	  \operatorname{sg}\!\left[\hat R_{\mathrm{Adv}}(q,h_k)\right]
	  \frac{1}{|h_k|}
	  \sum_{t=1}^{|h_k|}
	  \log \pi_\theta^{\mathrm{Adv}}(h_{k,t}\mid q,h_{k,<t}),
  \label{eq:adv-rb-loss}
\end{equation}
where $h_{k,t}$ denotes the $t$-th token of hint $h_k$, and
$\operatorname{sg}[\cdot]$ denotes stop-gradient. While REINFORCE updates often suffer from high variance and instability, our formulation helps mitigate this issue. Rather than being a single-shot noisy reward, the advantage $\hat R_{\mathrm{Adv}}(q,h_k)$ is derived from success rates averaged over $G_1$ and $G_3$ trajectories. This built-in variance reduction makes it feasible to treat the empirical degradation as a scalar advantage for updating the Adversary's generation probabilities. 

Alternatively, when sampling a sufficiently large batch of hint candidates per question, the Adversary can use a grouped GRPO-style update to normalize within-question credit assignment. Here, we focus on the more accessible and effective default configuration.
% Alternatively, when sampling a sufficiently large batch of hint candidates per question, the Adversary can use a grouped GRPO-style update with a shaped outcome score to mitigate sampling variance. To maintain narrative focus on Seirênes's most accessible and effective default configuration, we defer the full derivation of this high-budget extension to Appendix~\ref{app:adv-grpo}.

% When multiple hints per question are available, the Adversary update can instead
% use a grouped GRPO-style advantage over hint candidates.  Because the empirical
% degradation is a noisy finite-sample estimate rather than a direct binary
% verifier reward, applying group-centering directly to $\hat\Delta(q,h_k)$ can
% assign positive advantages to ineffective hints when sampling noise makes them
% appear relatively less harmful than other candidates.  We therefore compute a
% shaped outcome score $s(q,h_k)$ before group normalization, preserving
% fine-grained differences among harmful hints while mapping ineffective or
% helpful hints to a non-positive failed-attack region.

\noindent\textbf{Robustness branch.} Finally, to reinforce correct reasoning under misleading context, this branch applies GRPO to the R3 trajectories. Let $\mathcal{H}_q \subseteq \{h_k\}_{k=1}^{G_2}$ denote the hint subset selected for the update, and let $\mathcal{Q}_{\mathrm{rob}} = \{q : |\mathcal{H}_q| > 0\}$. The robustness objective is defined as:
\begin{equation}
  \mathcal{L}_{\mathrm{robust}}
  =
  \frac{1}{|\mathcal{Q}_{\mathrm{rob}}|}
  \sum_{q\in\mathcal{Q}_{\mathrm{rob}}}
  \frac{1}{|\mathcal{H}_q|}
  \sum_{h_k\in\mathcal{H}_q}
  \left[
    -J_{\mathrm{GRPO}}
    \!\left(\theta;\{(q\oplus h_k,o_{k,j}^{(3)},R_{k,j}^{(3)})\}_{j=1}^{G_3}\right)
  \right].
  \label{eq:robust-loss}
\end{equation}
The GRPO update is computed separately within each hint-conditioned group, while outer averages balance the signal across questions and their corresponding adversarial subsets.

\setlength{\textfloatsep}{5pt}
\begin{algorithm}[t]
\caption{Seirênes Training Orchestration}
\label{alg:Seirênes-training}
\begin{algorithmic}[1]
\REQUIRE Paired rollout bundle $\mathcal{D}$; pending queues $\{\mathcal{P}_s\}$;
flush sizes $\{M_s\}$, with $s\in\{c,\mathrm{adv},r\}$
\STATE Compute $\hat p(q)$, $\hat p(q,h_k)$, and
$\hat R_{\mathrm{Adv}}(q,h_k)$ from $\mathcal{D}$.
\STATE Build candidate batches $\mathcal{C}_s$ and keep
$\widetilde{\mathcal{C}}_s=\{g\in\mathcal{C}_s:\operatorname{A}_s(g)!=0\}$
for $s\in\{c,\mathrm{adv},r\}$.
\FOR{$s\in\{c,\mathrm{adv},r\}$}
  \STATE $\mathcal{P}_s \leftarrow
  \mathcal{P}_s \cup \widetilde{\mathcal{C}}_s$.
  \IF{$|\mathcal{P}_s|\ge M_s$}
    \STATE Update $\theta$ with $\mathcal{L}_s$ on $\mathcal{P}_s$; advance
    $\mathcal{P}_s$.
  \ENDIF
\ENDFOR
\end{algorithmic}
\end{algorithm}

\subsection{Training orchestration}
\label{sec:training-orchestration}
To preserve the distinct credit semantics of each role, Seirênes avoids naive batch pooling. Instead, it decouples the three branches into separate update streams, ensuring that gradients from clean reasoning do not interfere with those from adversarial generation or robustness recovery. Once a paired rollout bundle is evaluated, Seirênes constructs role-specific candidate batches. These candidates are then filtered by 
% $\operatorname{Signal}_s(g)>0$, which 
removing zero-advantage data ($\operatorname{A}_s(g)!=0$) from each stream.

The valid data is accumulated in stream-specific pending queues until the corresponding flush size $M_s$ is reached. As summarized in Algorithm~\ref{alg:Seirênes-training}, these queues act as cadence controllers: because clean, Adversary, and robustness groups survive filtering at different rates, the queues absorb these asymmetric arrival rates. An update to the shared policy is triggered only when a specific stream has gathered sufficient data to form a stable gradient estimate, after which the system advances the queue state. This orchestration ensures that each objective maintains its intended grouping while jointly optimizing the same underlying model. Long-horizon training dynamics and per-stream stability diagnostics are reported in Appendix~\ref{app:training-dynamics}.

% Because these branches optimize different trajectories under distinct credit
% assignments, their objectives are processed through separate update streams
% rather than aggregated into a single pooled batch.  Once a paired rollout bundle
% is generated and evaluated, Seirênes constructs role-specific candidate batches.
% It then filters out zero-signal groups, accumulates the valid data in
% stream-specific pending queues until a preset threshold (flush size) is reached,
% and applies the corresponding loss to update the shared policy.
% Algorithm~\ref{alg:Seirênes-training} summarizes this logical orchestration.

% In Algorithm~\ref{alg:Seirênes-training}, the index
% $s \in \{c, \mathrm{adv}, r\}$ denotes the clean, Adversary, and robustness
% streams, corresponding to the respective objectives
% $\mathcal{L}_{\mathrm{clean}}$, $\mathcal{L}_{\mathrm{Adv}}$, and
% $\mathcal{L}_{\mathrm{robust}}$ defined earlier.  The condition
% $\operatorname{Signal}_s(g) > 0$ retains only rollout groups with valid learning
% signals.  Crucially, these stream-specific queues act as cadence controllers. Because clean, Adversary, and robustness groups survive zero-signal filtering at different rates, the queues absorb these asymmetric arrival rates and trigger an update only when a stream reaches its designated group capacity $M_s$. This ensures that clean reasoning, adversarial hint generation, and hint-conditioned robustness preserve their distinct grouping and credit semantics while updating the same shared policy.

\subsection{Efficient implementation}
\label{sec:efficient-implementation}
Three-round evaluation multiplies rollout costs. We therefore introduce practical strategies to maintain adversarial integrity while rendering overhead tractable.

\noindent\textbf{Bounded FIFO buffers.} To avoid the ``reject-and-regenerate'' waste common in online DAPO-style filtering (where valid data at batch boundaries is often discarded), the pending queues are implemented as role-specific Bounded FIFO (first in, first out) buffers. When a stream $s$ reaches its flush size $M_s$, Seirênes consumes $M_s$ groups for an update but crucially does not flush the entire buffer. Any remaining valid groups are carried over to subsequent steps; if a buffer reaches its capacity, Seirênes temporarily suspends sampling for that stream until the next update consumes the backlog. This is the physical realization of the advanced operation in Algorithm~\ref{alg:Seirênes-training}. 

However, this persistence introduces data staleness as trajectories lag behind the latest policy. To mitigate off-policy drift, we discard stale rollout groups that lag significantly behind the current policy. This design differs from Selective Sample Replay~\cite{wang2025vl} by prioritizing temporal recency for policy–data alignment rather than advantage magnitude for signal density.

% The pending queues in Algorithm~\ref{alg:Seirênes-training} are implemented as role-specific FIFO buffers. When the buffer for stream $s$ reaches its flush threshold, Seirênes consumes $M_s$ groups for an actor update but crucially does not clear the entire buffer; any remaining valid groups are carried over across training steps to contribute to later updates. This avoids the reject-and-regenerate waste of purely online DAPO-style filtering, where valid leftover groups at batch boundaries are typically discarded. The inherent trade-off of this decoupled buffering is data staleness, as the trajectories consumed during an actor update may lag several steps behind the current policy. To manage this off-policy drift, we enforce a configured lag bound: any buffered group that becomes too stale relative to the active model is permanently evicted. Importantly, this systems-layer design differs fundamentally from selective sample replay methods, such as VL-Rethinker SSR, which store nonzero-advantage samples in a persistent replay buffer and sample them with probabilities proportional to advantage magnitude. Instead, our FIFO buffers---coupled with dynamic backpressure---serve purely to preserve unused valid rollouts, explicitly balance hardware utilization with off-policy drift, and prevent auxiliary streams from outpacing the clean backbone.

\noindent\textbf{Latency-aware rollout scheduling.} Although our exposition introduces $R_1$, $R_2$, and $R_3$ as sequential
stages for narrative clarity, only $R_3$ carries a true data dependency on
the outputs of $R_1$ and $R_2$: both $R_1$ and $R_2$ condition solely on the original question $q$
and impose no ordering constraint on each other. Seirênes exploits this slack
by preconstructing the $R_2$ prompts and submitting them concurrently with
the $R_1$ prompts to the rollout backend, collapsing the critical path to
$T_{\mathrm{merged}} = T_{12} + T_{R_3}$. This merging is particularly
effective because of the sequence length imbalance intrinsic to the RL loop:
$R_1$ and $R_3$ are long reasoning trajectories that, as response lengths
diverge within a batch, leave growing computational ``bubbles'' in the inference engine, whereas $R_2$ adversarial hints are markedly
shorter. With continuous batching, the brief $R_2$ sequences are scheduled into these idle
slots, so that the overhead of $R_2$ is effectively absorbed into the $R_1$
generation process and $T_{12}$ tightly approximates $T_{R_1}$. The same
schedule incidentally hides reward-computation latency: once $R_2$ hints are
available, $R_3$ rollouts proceed immediately, and the $R_1$ reward
verification is overlapped asynchronously with $R_3$ generation rather than
sitting on the critical path.

\noindent\textbf{Mastery-aware sampling.} 
As training progresses, certain problems become too easy for the Reasoner, solve cleanly even under adversarial hints. Continuing the full three-round loop on such instances yields diminishing learning signals while consuming a substantial rollout budget. We therefore maintain a mastered set $\mathcal{M}$ to deprioritize these questions in the active pool. While mastery-aware sampling is a general cost-control mechanism for RLVR, it is uniquely synergistic with our adversarial setting. By demanding robustness against targeted hints rather than mere clean-prompt success, it provides a more rigorous pruning signal, redirecting resources to instances that still expose model vulnerabilities. Formal operational criteria and the mastery indicator are detailed in Appendix~\ref{app:mastery}.

% ============================================================
%  Experiments
% ============================================================

\section{Experiments}
\label{sec:exp}

\subsection{Experimental setup}
\label{sec:exp-setup}

\noindent\textbf{Benchmarks.} We evaluate on seven mathematical reasoning benchmarks:
AIME 2024--2026~\cite{aime2024,aime2025,aime2026}, IMO-Bench~\cite{luong2025towards},
Minerva Math~\cite{lewkowycz2022solving}, OlympiadBench~\cite{he2024olympiadbench},
and HMMT 2026~\cite{balunovic_srimatharena_2025}.
For AIME 2024--2026 and HMMT 2026 we report average accuracy across 32 independent samples per problem (avg@32); for MINERVA Math, OlympiadBench,
and IMO-Bench we report avg@4.

\noindent\textbf{Baselines.} We compare against two main groups. \textit{Standard RL baselines} establish the frontier without auxiliary context, including DAPO~\cite{yu2025dapo}, ExGRPO~\cite{zhan2025exgrpo} and Dr.GRPO~\cite{liu2025understanding}. \textit{Hint-augmented baselines} inject auxiliary context into the RL loop with \emph{cooperative} intent: SAGE~\cite{liao2026self} and Scaf-GRPO~\cite{zhang2025scaf} use reference-guided
  scaffolds, LUFFY~\cite{yan2025learning} relies on off-policy traces from stronger external models, and InT~\cite{yang2026int} relies on external references to localize reasoning errors. For compute-controlled comparisons, we train Seirênes and the DAPO/Dr.GRPO baselines under the same training budget; for other baselines, we use the released checkpoints when available.

\noindent\textbf{Training and evaluation.}
We train Qwen2.5-7B-Instruct~\cite{qwen2.5}, Qwen3-4B-Instruct-2507~\cite{qwen3}, and Qwen3-30B-A3B-Instruct-2507~\cite{qwen3}. Training uses model-specific difficulty-balanced subsets from DAPO-Math-17K~\cite{yu2025dapo} and OpenR1-Math~\cite{openr1}. Unless otherwise stated, Seirênes uses $G_1{=}G_3{=}8$ Reasoner rollouts and $G_2{=}2$ adversarial hints per prompt; all evaluations use temperature $1.0$ and top-p $1.0$. Full data-construction and optimization details are in Appendix~\ref{app:exp-setup-details}.

\vspace{-2mm}

\begin{table}[t]
    \vspace{-2mm}
  \centering
  \caption{%
    Results on mathematical reasoning benchmarks.
    $\boldsymbol{\Delta}$ is the absolute improvement over the instruction-tuned baseline of the same model family.
    \textsuperscript{\dag}~Trained under the same budget as Seirênes.
  }
  \label{tab:math-main}
  \resizebox{\linewidth}{!}{%
  \begin{tabular}{lccccccccr}
  \toprule
  \textbf{Model} & \textbf{AIME24} & \textbf{AIME25} & \textbf{AIME26} & \textbf{IMO-Bench} & \textbf{MINERVA} & \textbf{Olympiad} & \textbf{HMMT26} & \textbf{AVG} & $\boldsymbol{\Delta}$\textbf{AVG} \\
  \midrule
  \textit{Qwen2.5-7B-Instruct}        & 10.2 & 5.5  & 5.3  & 8.8  & 29.4 & 34.9 & 2.7  & 13.8 & -- \\
  \quad ExGRPO~\cite{zhan2025exgrpo}                         & 14.2 & 13.2 & 9.0  & 9.9  & 29.9 & 42.0 & 6.0  & 17.7 & +3.9 \\
  \quad LUFFY~\cite{yan2025learning}      & 13.9 & 12.9 & 12.3 & 11.0 & 32.8 & 41.5 & 6.3  & 18.7 & +4.9 \\
  \quad SAGE~\cite{liao2026self}       & 15.8 & 13.7 & 9.3  & 11.2 & 31.6 & 42.2 & 6.3  & 18.6 & +4.8 \\
  \quad DAPO\textsuperscript{\dag}~\cite{yu2025dapo} & 14.1 & 12.0 & 9.3 & 9.5 & 30.5 & 42.1 & 4.3 & 17.4 & +3.6 \\
  \quad \textbf{Seirênes (Ours)\textsuperscript{\dag}} & \textbf{22.8} & \textbf{15.4} & \textbf{17.8} & \textbf{13.3} & \textbf{34.7} & \textbf{48.0} & \textbf{8.5} & \textbf{22.9} & \textbf{+9.1} \\
  \midrule
  \textit{Qwen3-4B-Instruct}          & 61.5 & 48.8 & 56.8 & 34.1 & 35.4 & 66.3 & 31.9 & 47.8 & -- \\
  \quad Scaf-GRPO~\cite{zhang2025scaf} & 61.6 & 49.1 & 54.3 & 33.9 & 38.6 & 67.1 & 32.2 & 48.1 & +0.3 \\
  \quad SAGE~\cite{liao2026self}       & 60.1 & 49.1 & 57.8 & 32.5 & 41.1 & 67.2 & 29.9 & 48.2 & +0.4 \\
  \quad InT~\cite{yang2026int}         & 66.6 & 58.9 & 64.7 & 38.8 & 38.3 & 67.8 & 37.9 & 53.3 & +5.5 \\
  \quad Dr.GRPO\textsuperscript{\dag}~\cite{liu2025understanding} & 65.0 & 63.3 & 64.9 & 38.0 & 39.4 & 68.3 & 37.2 & 53.7 & +5.9 \\
  \quad DAPO\textsuperscript{\dag}~\cite{yu2025dapo} & 63.0 & 60.2 & 62.2 & 37.7 & 40.1 & 68.3 & 33.0 & 52.1 & +4.3 \\
  \quad \textbf{Seirênes (Ours)\textsuperscript{\dag}} & \textbf{74.3} & \textbf{64.9} & \textbf{70.8} & \textbf{41.9} & \textbf{42.8} & \textbf{71.9} & \textbf{39.4} & \textbf{58.0} & \textbf{+10.2} \\
  \midrule
  \textit{Qwen3-30B-A3B}              & 75.0 & 60.6 & 75.1 & 40.5 & 35.0 & 70.0 & 41.0 & 56.7 & -- \\
  \quad DAPO\textsuperscript{\dag}~\cite{yu2025dapo} & 78.6 & 66.8 & 75.2 & 44.3 & 40.6 & 72.8 & 42.3 & 60.1 & +3.4 \\
  \quad \textbf{Seirênes (Ours)\textsuperscript{\dag}} & \textbf{81.8} & \textbf{74.1} & \textbf{79.7} & \textbf{46.7} & \textbf{41.7} & \textbf{74.3} & \textbf{49.0} & \textbf{63.9} & \textbf{+7.2} \\
  \bottomrule
  \end{tabular}%
  }
  \end{table}

\subsection{Main results}
\label{sec:exp-math}
Tab.~\ref{tab:math-main} presents the clean, no-hint mathematical reasoning performance across three backbones. This evaluation tests whether the Seirênes framework---co-evolving a single policy as both Adversary and Reasoner---translates into stronger standalone reasoning ability in the base models. Across backbones, Seirênes improves upon the instruction-tuned baselines, showing positive gains across all evaluated datasets and raising the average accuracy. We highlight three patterns behind these results.

First, the improvements are not confined to weaker models. Qwen3-4B-Instruct
starts from a higher baseline than Qwen2.5-7B-Instruct
(47.8 vs.\ 13.8 AVG) yet achieves a higher gain (+10.2 vs.\ +9.1). 
This suggests that the adversarial loop remains useful beyond low-capacity initializations. The pattern
also holds on a different architecture: on Qwen3-30B-A3B-Instruct,
a sparse MoE backbone where RL can be less stable, Seirênes increases performance 
across the evaluation suite and outperforms the DAPO baseline by
+3.8 AVG. Furthermore, the gains extend to benchmarks released
after the base models' training cutoff (AIME 2026, HMMT 2026), where
template memorization is less plausible.

Second, the gains persist across benchmarks that stress the models in different ways. Seirênes improves recent high-difficulty benchmarks such as AIME 2026 and HMMT 2026, while also improving broader mathematical reasoning evaluations such as IMO-Bench, MINERVA, and OlympiadBench. The effect is also visible on splits where the base model initially performs poorly: on Qwen2.5-7B-Instruct, HMMT 2026 rises from 2.7 to 8.5 and AIME 2026 from 5.3 to 17.8; on Qwen3-4B-Instruct, IMO-Bench moves from 34.1 to 41.9 and MINERVA from 35.4 to 42.8. Thus, the aggregate gains are not driven solely by benchmarks where the base model is already strong, but also appear on benchmarks where the base model has the most room to improve.

Third, cooperative-hint methods (LUFFY, SAGE, Scaf-GRPO, InT) show
that auxiliary context is a productive ingredient for RLVR. Seirênes
demonstrates competitive results using context in the opposite
direction---adversarial rather than cooperative. This suggests that
misleading context can serve as a training signal comparable to helpful
scaffolding and that the adversarial loop itself is a viable axis for
model improvement. Finally, the same-budget comparisons with DAPO and
Dr.GRPO indicate that these gains are not simply recovered
by allocating more compute to standard rollouts.

\begin{figure}[t]
\vspace{-2mm}
  \centering
  \begin{minipage}[t]{0.495\linewidth}
    \centering
    \includegraphics[width=\linewidth]{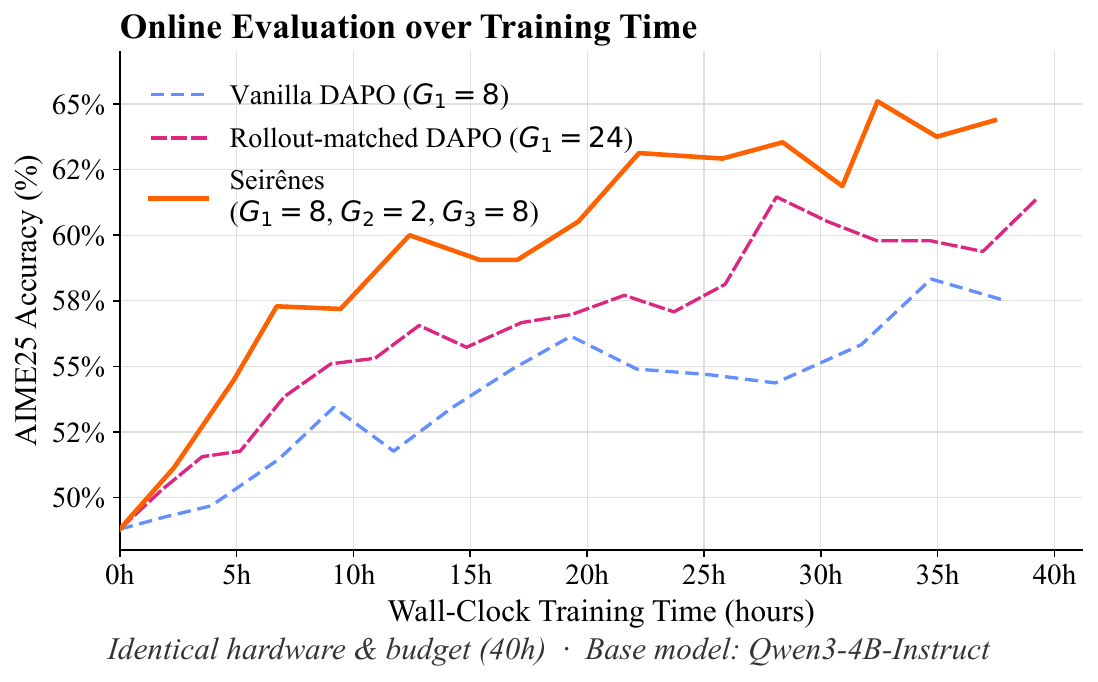}
  \end{minipage}
  \hfill
  \begin{minipage}[t]{0.495\linewidth}
    \centering
    \includegraphics[width=\linewidth]{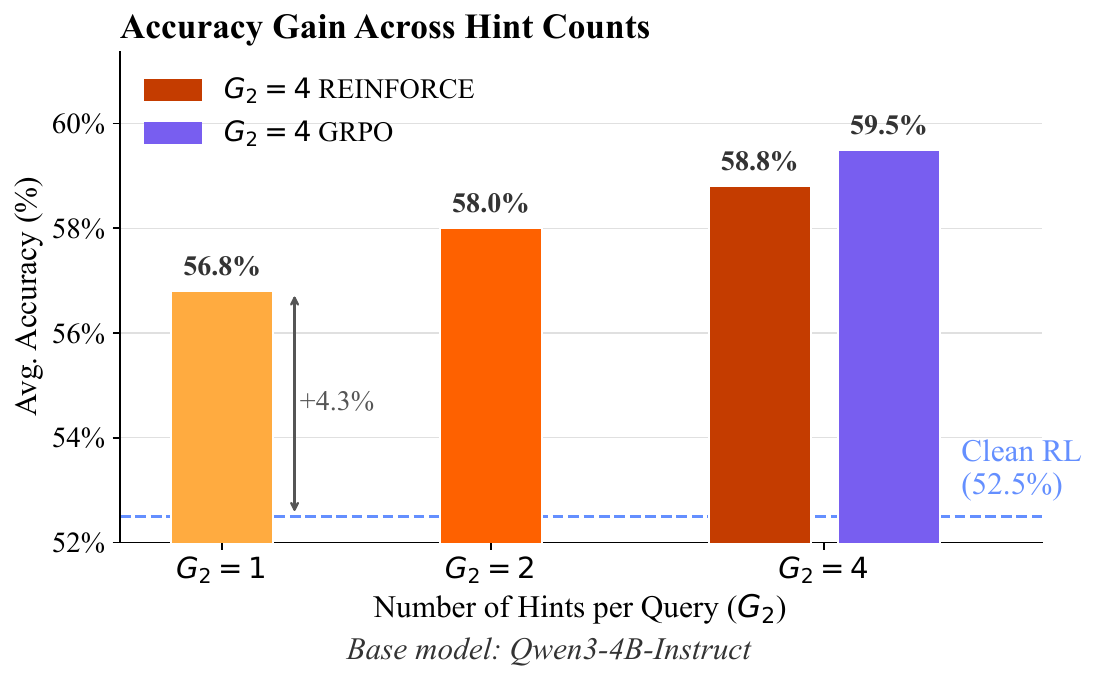}
  \end{minipage}
  \vspace{-4mm}
  \caption{%
    (a) AIME25 accuracy vs.\ wall-clock training time for three budget-matched conditions on identical hardware.
    (b) AVG accuracy vs.\ number of adversarial hints $G_2$ per step.
  }
  \label{fig:p03-curves}
  \vspace{-4mm}
\end{figure}

\begin{figure}[t]
  \centering
  \begin{minipage}[t]{0.495\linewidth}
    \centering
    \includegraphics[width=\linewidth]{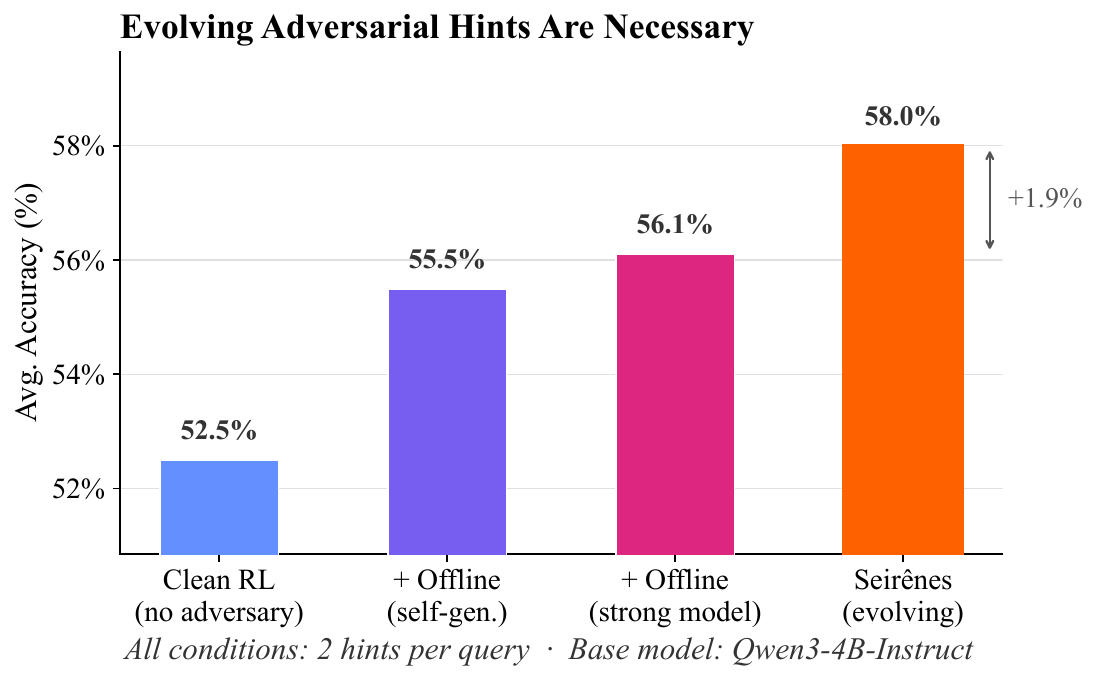}
  \end{minipage}
  \hfill
  \begin{minipage}[t]{0.495\linewidth}
    \centering
    \includegraphics[width=\linewidth]{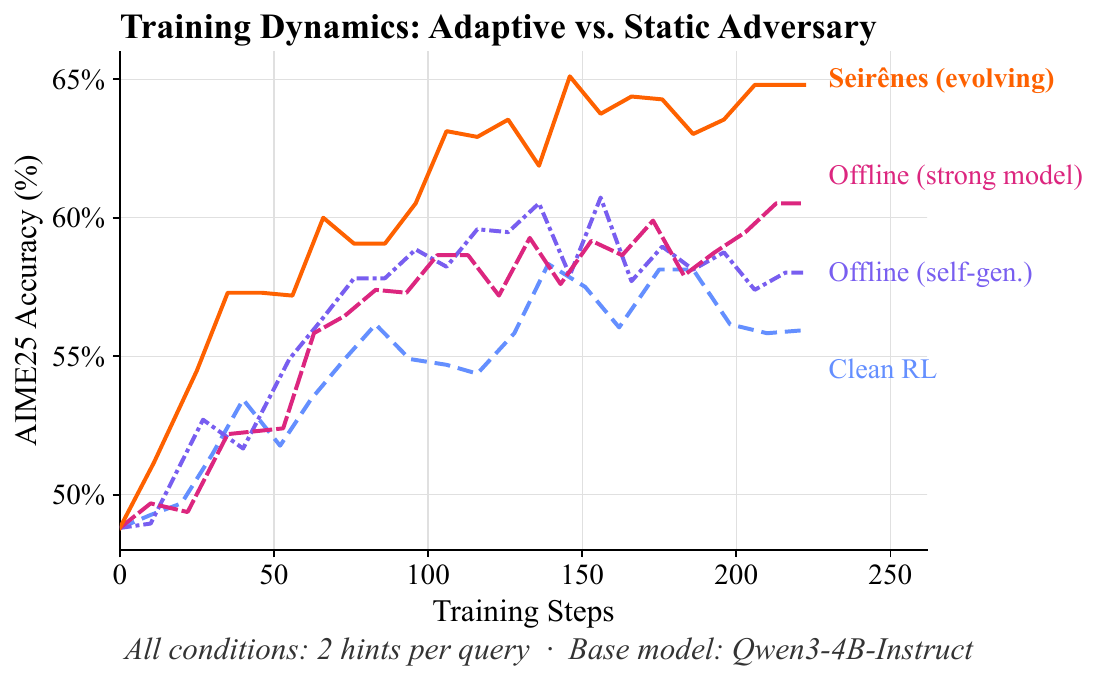}
  \end{minipage}
   \vspace{-4mm}
  \caption{%     
    \textbf{Static vs.\ evolving adversarial hints.}                                           
    (a) AVG accuracy across benchmarks for each condition.
    (b) AIME25 online accuracy over training steps.
  }
  \label{fig:adaptive-necessary}
\end{figure}
\vspace{-2mm}
\subsection{Analysis}
\label{sec:exp-analysis}

\noindent\textbf{Budget control.} A natural concern is that Seirênes's gains might benefit from extra per-step
compute introduced by R2 and R3 rollouts. We thus ablate three configurations on identical hardware under a fixed 40-hour wall-clock budget:
vanilla DAPO with $G_1{=}8$ clean rollouts, rollout-matched DAPO with
$G_1{=}24$ clean rollouts, and Seirênes with $G_1{=}8$, $G_2{=}2$, and $G_3{=}8$.
The comparison isolates the source of the additional Reasoner rollouts: more
clean rollouts versus $G_2G_3$ hint-conditioned rollouts driven by the
Adversary. Fig.~\ref{fig:p03-curves}(a) plots AIME25 accuracy against
wall-clock time.

Under this controlled comparison, Seirênes improves more rapidly throughout the
40-hour window. At matched wall-clock points, it reaches higher AIME25 accuracy
earlier, indicating that the adversarially conditioned rollouts provide an
effective training signal within the same compute envelope. Its endpoint is also
the highest among the tested configurations, and the curve remains upward-moving
at the end of the budget window. These trends support the view that Seirênes turns the
additional rollout budget into faster and higher clean-reasoning improvement
through adversarial co-evolution.

\noindent\textbf{Effect of hint count.} We vary the adversarial hint count ($G_2\in\{1,2,4\}$, Fig.~\ref{fig:p03-curves}(b)) to test whether additional misleading contexts further improve training. Compared to a budget-matched baseline that omits the adversarial $R_2$ and $R_3$ rollouts, introducing even a single hint yields a substantial performance jump. Our default setting with $G_2{=}2$ further improves average accuracy to $58.0\%$. Scaling to $G_2{=}4$ achieves the highest absolute performance, indicating that the framework can continuously unlock further reasoning advantages when allocated a larger compute budget. While this trend suggests strong scaling potential, deploying $G_2 \ge 4$ until convergence significantly inflates the per-step inference cost and currently exceeds our computational resources. Fully exploring the upper bounds of this adversarial scaling remains a direction for future work.

\noindent\textbf{Static vs. evolving adversarial hints.} Beyond the mere volume of misleading contexts, we investigate whether static hints generated offline can reproduce the observed performance gains. To isolate this, we evaluate four configurations on the Qwen3-4B-Instruct (Fig.~\ref{fig:adaptive-necessary}): standard clean RL, two static hint baselines, and Seirênes's evolving Adversary. Specifically, we evaluate \emph{static self-generated hints} (produced offline by the Qwen3-4B-Instruct backbone itself using the identical adversarial prompt) and \emph{static strong-model hints} (produced offline by Gemini~3~Flash~\cite{gemini3flash}). 

As illustrated in Fig.~\ref{fig:adaptive-necessary}, while static hints provide a useful learning signal, they fail to match the efficacy of the adaptive loop. Notably, even the high-quality offline hints generated by Gemini~3~Flash plateau below Seirênes in overall accuracy. This contrast suggests that an adversarial signal is most effective when it dynamically tracks the Reasoner's evolving capabilities. Because static hints act as fixed snapshots of earlier distractors, they tend to become less effective at exposing the model's current failure modes once the Reasoner learns to bypass them. Seirênes, by refreshing the adversarial distribution online, keeps misleading contexts better aligned with the current decision boundary. Appendix~\ref{app:evolve_necessary} quantifies this effect with per-step attack-strength statistics and clean-validation traces. Empirically, dynamic co-evolution averts early saturation, yielding further performance gains.

\noindent\textbf{Robustness to adversarial perturbations.} We next investigate whether training against evolving adversarial context transfers beyond the training hint distribution by evaluating robustness under structural math perturbations and generated distractors across Math-Perturb~\cite{huang2025math}, GSMIR~\cite{jiang2024enhancing}, MMLU-Perturb~\cite{wang2025adaptive}, and OpenBookQA-Perturb~\cite{wang2025adaptive}; benchmark details are deferred to Appendix~\ref{app:robustness-details}.

Tab.~\ref{tab:robustness} demonstrates Seirênes's robustness across diverse perturbation families. On MATH-Perturb, Seirênes achieves the highest accuracy on both splits, improving the average perturbation accuracy from 83.9 for the instruction-tuned model to 88.2. The robustness-degradation columns further show that Seirênes is less easily distracted by generated irrelevant contexts, obtaining the lowest average degradation among all methods. Interestingly, the cooperative hint-based baselines all exhibit larger average degradation than the instruction-tuned model. While this does not establish a causal mechanism, it is consistent with the possibility that training with helpful auxiliary context can make the model more willing to trust externally provided cues, whereas Seirênes trains the model to scrutinize such cues before incorporating them into its reasoning. Together, these results suggest that adversarial-context training improves robustness in a way that is not limited to the specific negative-hint distribution used during training. Rather than merely reinforcing memorized templates or shallow pattern shortcuts, Seirênes appears to encourage reasoning behavior that better survives both structural reformulations and distracting contextual evidence.

\begin{table}[t]
\vspace{-2mm}
\centering
\caption{%
  Robustness to adversarial perturbations.
  Best per column in \textbf{bold}.
}
\label{tab:robustness}
\resizebox{\linewidth}{!}{%
\begin{tabular}{@{}lccc cccc@{}}
\toprule
 & \multicolumn{3}{c}{\textbf{Perturb Acc} ($\uparrow$)}
 & \multicolumn{4}{c}{\textbf{Robustness Degradation} ($\downarrow$)} \\
\cmidrule(lr){2-4}\cmidrule(lr){5-8}
\textbf{Model}
  & \textbf{Math-P-Hard}
  & \textbf{Math-P-Simple}
  & \textbf{Avg}
  & \textbf{GSMIR}
  & \textbf{MMLU-P}
  & \textbf{OBook-P}
  & \textbf{Avg} \\
\midrule[0.3pt]
Qwen3-4B-Instruct          & 73.9          & 93.9          & 83.9          & \textbf{2.96}          & 9.39          & 19.03          & 10.46          \\[-1pt]
\quad Scaf-GRPO~\cite{zhang2025scaf}             & 71.7          & 93.9          & 82.8          & 3.57          & 9.12          & 19.03          & 10.57          \\[-1pt]
\quad SAGE~\cite{liao2026self}                  & 76.3          & 95.7          & 86.0          & 4.34          & 7.54          & 19.62          & 10.50          \\[-1pt]
\quad InT~\cite{yang2026int}                   & 76.1          & 94.8          & 85.4          & 7.39          & 8.25          & 17.92          & 11.19          \\[-1pt]
\quad \textbf{Seirênes (Ours)} & \textbf{79.1} & \textbf{97.4} & \textbf{88.2} & 3.05 & \textbf{4.91} & \textbf{16.21} & \textbf{8.06}  \\
\bottomrule
\end{tabular}%
}
\end{table}

% \noindent\textbf{What does the Adversary learn to say?} 
% Representative examples and diagnostics in Appendix~\ref{app:hint-effectiveness} show that the learned hints typically take the form of \emph{half-true} mathematical continuations, targeting plausible reasoning shortcuts rather than random noise.

\noindent\textbf{What does the Adversary learn to say?}
Representative examples and diagnostics in Appendix~\ref{app:hint-effectiveness} show that the learned hints typically take the form of \emph{half-true} mathematical continuations, exploiting plausible reasoning shortcuts rather than injecting random noise. Moreover, the hints generated by our 4B adversary 
% transfer beyond the training family and scale, 
even significantly degrade top-tier closed-source models (both Gemini and GPT), demonstrating the training pipeline's effectiveness and Seirênes' robustness; see Table~\ref{tab:transfer-attack} in Appendix.

\section{Conclusion}
We introduce Seirênes, a shared-parameter self-play RL framework that turns contextual interference into an internal training signal for mathematical reasoning. A single policy co-evolves as both an Adversary that generates plausible but distracting contexts and a Reasoner that recovers the core problem logic under such interference. Across model scales and seven mathematical benchmarks, Seirênes yields consistent gains for standalone reasoning ability.

\clearpage
{\small
\bibliographystyle{unsrtnat}
\bibliography{main}
}

\clearpage
\appendix
\section{Extended related work}
\label{app:related-extended}
\noindent\textbf{Adversarial examples for language understanding.} Text-level adversarial inputs are a long-standing tool for exposing brittleness in neural NLP models. Jia and Liang~\cite{jia2017adversarial} show that appending semantically related but task-irrelevant sentences to reading-comprehension passages causes sharp accuracy drops, a phenomenon directly mirrored at the LLM scale by GSM-IC~\cite{shi2023large}, which demonstrates that grade-school math problems can be derailed simply by inserting irrelevant quantitative sentences. Wallace et al.~\cite{wallace2019universal} further show that short, input-agnostic trigger strings transfer across both inputs and target models, foreshadowing the cross-model attack pattern we observe in Section~\ref{app:hint-effectiveness}, where a 4B Seirênes Adversary degrades GPT-5.1~\cite{gpt5_1} and Gemini-3-Flash~\cite{gemini3flash}. These works share Seirênes's core threat model of using plausible text to derail reasoning, yet they relegate adversarial context strictly to evaluation, testing static reasoners against fixed attacks. Seirênes fundamentally inverts this approach. Instead of treating adversarial vulnerability as a static metric, it integrates the attacker into the RL loop, co-evolving the Reasoner against a dynamic Adversary to turn a traditional vulnerability into a constructive training signal.

\noindent\textbf{Adversarial training and co-evolving RL adversaries.} The constructive use of adversaries during training finds its canonical form in Madry et al.~\cite{madry2017towards}, which formalizes adversarial training as a saddle-point minimax optimization with an inner PGD attacker producing norm-bounded input perturbations. Pinto et al.~\cite{pinto2017robust} (RARL) lift this concept to reinforcement learning by training a protagonist policy alongside a disturbance adversary that perturbs environment dynamics. PAIRED~\cite{dennis2020emergent} pushes this further by having an adversary generate regret-maximizing environments, creating an autocurriculum that continuously challenges the agent at its capability frontier. Seirênes inherits this minimax lineage but introduces three substantive shifts. \emph{First}, the perturbation takes the form of a coherent natural-language hint rather than a norm-bounded token manipulation or a scalar control disturbance. This shifts the attack surface from gradient-based token optimization to the semantic reasoning level. \emph{Second}, the core components of the environment, including the question, the verifier, and the action space, remain strictly unchanged. Only the input \emph{context} is perturbed, ensuring the task remains well-specified while still posing a rigorous challenge. \emph{Third}, the Adversary and the Reasoner share parameters and are differentiated solely by role conditioning. This design collapses the separate networks typically required in RARL or PAIRED into a single update flow, avoiding a twofold increase in parameter count.

\noindent\textbf{LLM red-teaming and shared-parameter self-play.} A parallel line of work uses adversarial probing to characterize and improve safety-aligned LLMs. Perez et al.~\cite{perez2022red} and Ganguli et al.~\cite{ganguli2022red} prompt or train language models to attack other language models, surfacing harmful behaviors at scale; GCG~\cite{zou2023universal} and PAIR~\cite{chao2025jailbreaking} construct adversarial prompts via gradient search and iterative black-box querying, respectively, to bypass safety alignment. Constitutional AI~\cite{bai2022constitutional} and SPIN~\cite{chen2024self} further illustrate that a single language model can play two roles via shared parameters---critic versus actor in the former, and weak-data-generator versus stronger-target in the latter. Seirênes adopts the shared-parameter trick from this lineage but differs in two essential respects. The optimization target is \emph{reasoning capability} on a verifiable task rather than safety alignment or weak-to-strong distillation, so the adversarial signal must take the form of mathematically plausible derivations rather than alignment-bypassing prompts. And the use of attacks is fully in-loop: adversarial hints are produced by the same policy being optimized and consumed within the same RL update, rather than serving as one-shot evaluation probes~\cite{perez2022red, zou2023universal, chao2025jailbreaking}, a fixed dataset for downstream supervised tuning~\cite{ganguli2022red}, or a cooperative critic-actor arrangement aimed at distributional alignment~\cite{bai2022constitutional, chen2024self}. To our knowledge, Seirênes is the first framework to combine all three properties---a learned natural-language attacker, a shared-parameter self-play loop, and a verifiable reasoning objective---under a single online RL update.

\noindent\textbf{Reasoning Robustness under Contextual Perturbations.} Despite the striking capability gains driven by RLVR and self-evolving training, reasoning LLMs often exhibit     
  brittleness when operating outside the pristine conditions of standard benchmarks. Recent evaluations including   
  MATH-Perturb, RUPBench, and Math-RoB report severe degradation under structural reformulations, semantic          
  perturbations, and incomplete information~\cite{huang2025math,wang2024rupbench,yu2025benchmarking}. Concurrently, 
  complementary analyses highlight memorization, format sensitivity, and a reliance on superficial cues as          
  persistent confounders in assessing true reasoning capabilities~\cite{xie2025memorization,mousavi2026garbage}.    
  Beyond structural shifts, contextual distraction poses another significant challenge: biasing features can        
  misdirect chain-of-thought rationale~\cite{turpin2023language}, while task-irrelevant but semantically related
  contexts severely impair reasoning performance~\cite{wang2025adaptive,wu2024easily}. These vulnerabilities
  frequently surface in real-world deployment scenarios plagued by noisy retrieval documents, hard distractors, and
  erroneous tool outputs~\cite{Chen2023BenchmarkingLL,amiraz2025distracting,lee2026lost}.

\section{Experimental reproducibility details}
\label{app:exp-setup-details}

\noindent\textbf{Training backbones and infrastructure.}
We train on three backbones of varying scale: Qwen2.5-7B-Instruct~\cite{qwen2.5}, Qwen3-4B-Instruct-2507~\cite{qwen3}, and Qwen3-30B-A3B-Instruct-2507~\cite{qwen3}. All models are trained using verl~\cite{sheng2024hybridflow} with vLLM~\cite{kwon2023efficient} for rollout generation. The Adversary and Reasoner share a single set of parameters and are separated only through role-conditioned prompting. The prompt template is shown in Appendix~\ref{app:prompt}.

\noindent\textbf{Training data construction.}
For training data, we pool problems from DAPO-Math-17K~\cite{yu2025dapo} and OpenR1-Math~\cite{openr1}. To ensure that the RL process is supported by a well-distributed dataset optimally aligned with each model's individual capacity, we construct a tailored subset for every backbone. Concretely, for each backbone, we run an offline pass with 8 rollouts per problem at temperature 1.0, use the empirical pass rate to estimate difficulty, and then perform balanced sampling across difficulty buckets.

\noindent\textbf{Optimization hyperparameters.}
We use AdamW with a constant learning rate of $1{\times}10^{-6}$, no warm-up, KL penalty disabled ($\beta{=}0$), and asymmetric clipping with $\varepsilon_{\mathrm{low}}{=}0.2$ and $\varepsilon_{\mathrm{high}}{=}0.28$~\cite{yu2025dapo}. Each rollout batch contains $256$ prompts with $G_1{=}G_3{=}8$ responses per prompt and a maximum response length of $16{,}384$ tokens. By default, the Adversary generates $G_2{=}2$ misleading hints per prompt; we ablate this choice in Section~\ref{sec:exp-analysis}. All evaluation inference uses temperature $1.0$ with top-p $1.0$.

\noindent\textbf{Stream scheduling and auxiliary losses.}
For all main runs, we set the stream flush sizes to
$M_c{=}\mathrm{128}$ clean prompt groups ($\mathrm{128} \times \mathrm{8}{=}\mathrm{1024}$ trajectories), $M_{\mathrm{adv}}{=}\mathrm{256}$
R2 adversarial hint trajectories, and $M_r{=}\mathrm{128}$ 
hint-conditioned Reasoner groups ($\mathrm{1024}$ trajectories). Stale buffered entries are evicted
after at most $\mathrm{3}$ optimizer-step lag for the clean stream,
$\mathrm{3}$ optimizer-step lag for the Adversary stream, and $\mathrm{3}$
optimizer-step lag for the robustness stream. The R2 adversarial hint
generation is capped at $\mathrm{2048}$ tokens.

\noindent\textbf{Training length.}
For the main results in Tab.~\ref{tab:math-main}, all three backbones use the
same nominal training schedule of $200$ rollout-collection steps. Because
Seirênes uses stream-specific FIFO buffers and branch-specific flush cadences,
the actual number of steps can fluctuate slightly around this
nominal step count across backbones and streams.

\section{Robustness evaluation details}
\label{app:robustness-details}

\noindent\textbf{Benchmark families.}
We evaluate robustness beyond the training hint distribution using two complementary perturbation families. The first family tests robustness to structural reformulations of math problems. MATH-Perturb-Simple and MATH-Perturb-Hard modify competition-level math problems through structural and numerical perturbations~\cite{huang2025math}, testing whether the model can preserve mathematical reasoning when the problem formulation changes. Because these splits remain math-domain evaluations, we report absolute perturbed accuracy.

The second family tests robustness to generated distractors. GSMIR~\cite{jiang2024enhancing} augments GSM8K-style~\cite{cobbe2021training} elementary math word problems with semantically related but irrelevant sentences. MMLU-Perturb~\cite{wang2025adaptive} and OpenbookQA-Perturb~\cite{wang2025adaptive} provide out-of-domain evaluations by injecting answer-preserving distractors into multiple-choice problems. MMLU~\cite{hendrycks2020measuring} spans broad academic subjects across STEM, humanities, and social sciences, while OpenbookQA~\cite{mihaylov2018can} requires applying elementary-science facts together with commonsense reasoning.

\noindent\textbf{Perturbation generation and metrics.}
For all benchmarks involving generated distractors, we use Gemini~3~Flash to produce the perturbing contexts. The prompts used for test-time perturbation generation are independent of the prompts used during Seirênes training. For GSMIR, MMLU-Perturb, and OpenbookQA-Perturb, we report robustness degradation rather than absolute perturbed accuracy, where lower degradation indicates that the model is less affected by the added irrelevant context.

\begin{figure*}[t]
  \centering
  \includegraphics[width=\linewidth]{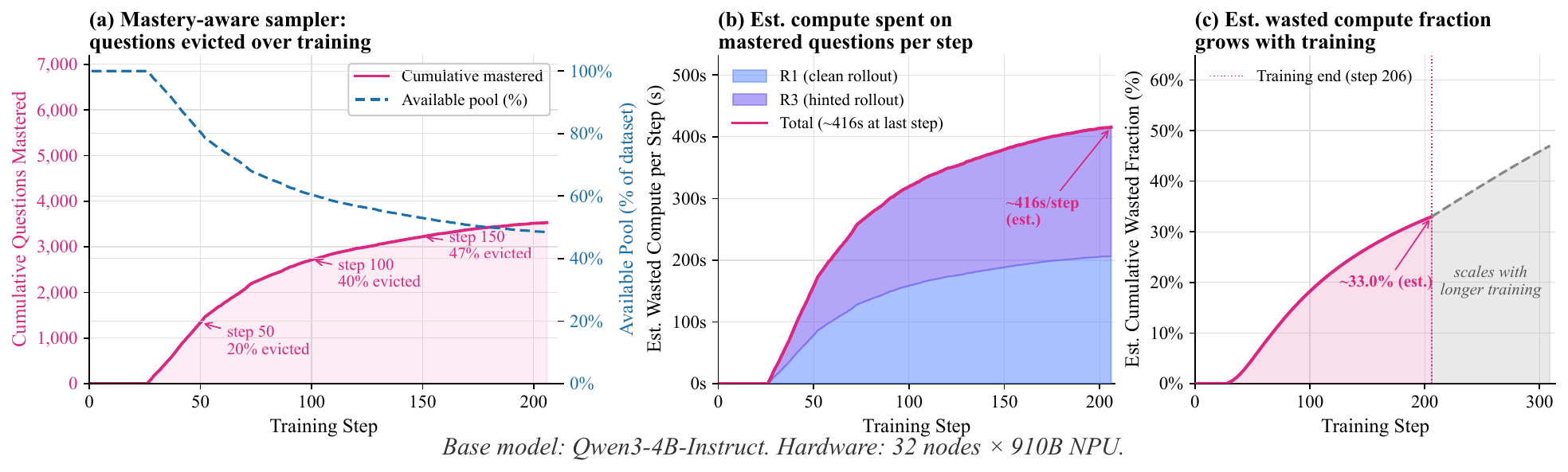}
  \caption{%
    \textbf{Mastery-aware sampler: eviction dynamics and estimated compute savings}
    (Qwen3-4B-Instruct, $|\mathcal{D}|{=}6{,}853$).
    \textbf{(a)}~Cumulative mastered set (left axis) and live adversarial pool fraction
    (right axis) over training.
    \textbf{(b)}~Per-step compute estimated to be spent on already-mastered questions,
    broken down by R1 (clean) and R3 (hinted) rollouts.
    \textbf{(c)}~Cumulative wasted fraction over training, with a logistic extrapolation
    (dashed; fit on observed steps 1--206) illustrating how the fraction grows with horizon.
    All estimates use a linear upper-bound model; actual savings are strictly smaller
    because rollout batch time is dominated by the longest trajectory.
  }
  \label{fig:mastery-combined}
\end{figure*}

\section{Mastery-aware sampling}
\label{app:mastery}

\noindent\textbf{Motivation and mechanism.}
As training progresses, some questions become too easy for the current Reasoner: it solves them cleanly, and the current Adversary's evaluated hints no longer induce failures. Continuing to run the full R1--R3 loop on such questions consumes rollout budget while producing little useful learning signal. The mastery-aware sampler therefore maintains an operational mastered set $\mathcal{M}$ and deprioritizes these questions in the live adversarial pool.

At evaluation step $\tau$, define
\begin{equation}
  m_\tau(q)
  =
  \mathbf{1}\!\left[
    \hat p_\tau(q)=1
    \;\wedge\;
    \forall h_k\in\mathcal{H}^{(\tau)}_q,\;
    \hat p_\tau(q,h_k)=1
  \right],
  \label{eq:mastery-indicator}
\end{equation}
where $\hat p_\tau(q)$ and $\hat p_\tau(q,h_k)$ denote the
estimators in Eqs.~\ref{eq:r1-clean-estimate} and~\ref{eq:r3-hinted-estimate}
computed from the rollouts collected at step $\tau$, $\mathcal{H}^{(\tau)}_q$ is the set of hinted groups for question $q$ that survive the zero-signal filter (Section~\ref{sec:training-orchestration}) at step $\tau$, and we evaluate $m_\tau(q)$ at every training step using the rollout statistics produced at that step. This is an operational criterion, not a claim of permanent mastery: $q$ enters $\mathcal{M}$ only once $m_\tau(q)=1$ holds for $K_m$ consecutive evaluation steps (where $K_m=1$ in our experiment). The aggressive choice $K_m=1$ is empirically reliable because the joint all-correct condition over $G_1$ clean and $\sum_k G_3$ hinted rollouts already provides strong implicit averaging at a single step; the post-hoc audit in Tab.~\ref{tab:mastery-verify} confirms that the resulting retired set has $98.9\%$ mean@8 with no question below $4/8$. While mastery-aware sampling can be viewed as a general cost-control mechanism for RLVR training, it is especially aligned with our adversarial setting because retirement requires robustness to targeted hints rather than clean-prompt success alone.

\noindent\textbf{What does mastery-aware sampling actually buy us?}
The practical value of mastery-aware sampling stems from two coupled effects: computational efficiency and eviction reliability. First, because Seirênes's three-round paired rollout incurs a per-question overhead, retiring robust questions directly reduces wasted compute. Second, because eviction relies on success under adversarial interference, the sampler should ideally retire only genuinely mastered questions. We evaluate the sampler along these two axes.

To quantify compute savings, we track the cumulative mastered set on the 6,853-question Qwen3-4B-Instruct pool. We estimate the theoretical saved rollout time via a linear projection, assuming rollout work scales proportionally with the active pool size:
\[
  \widehat{T}_{\text{saved}}(s)
  \;=\;
  \tfrac{|\mathcal{M}_s|}{|\mathcal{D}|}
  \cdot \bigl(\bar T_{R1}+G_2\bar T_{R3}\bigr),
\]
where $\bar T_{R1}{\approx}401\,\mathrm{s}$ and
$\bar T_{R3}{\approx}406\,\mathrm{s}$ are the average rollout
times measured in a $G_2{=}1$ probe run on the same backbone and
infrastructure. The reported percentage saving is invariant to $G_2$ because
both the retired and active per-step compute scale by the same factor
$\bar T_{R1}+G_2\,\bar T_{R3}$; we therefore use these $G_2{=}1$ timings as
a representative scale. 

To assess eviction reliability, we conduct a post-hoc audit by re-evaluating each retired question with $n=8$ clean rollouts at temperature 1.0. We compare this accuracy distribution against a matched DAPO sampler keyed on clean-success retirement. Both samplers share identical $K_m$ requirements, evaluation cadence, and step horizon. The only difference is that its mastery indicator weakens our criterion $\hat p_\tau(q){=}1\,\wedge\,\forall h_k\!\in\!\mathcal{H}^{(\tau)}_q,\,\hat p_\tau(q,h_k){=}1$ to clean-success alone, $\hat p_\tau(q){=}1$.

\noindent\textbf{Results.}
The cumulative mastered set grows rapidly: 1,339 questions (20\%) by step 50, 2,713 questions (40\%) by step 100, and 3,531 questions (52\%) by step 206 (Fig.~\ref{fig:mastery-combined}). Under the linear projection, the sampler reclaims approximately $33.9\%$ of the total training compute, with the per-step reduction reaching $52\%$ by the end of training.

Post-hoc verification confirms that our hint-conditioned threshold identifies a higher-confidence mastered set than clean-success retirement alone. As shown in Tab.~\ref{tab:mastery-verify}, Seirênes's retired questions achieve $98.9\%$ mean@8, compared to $95.5\%$ for the matched DAPO sampler. They also obtain a higher all-correct rate ($93.0\%$ vs.\ $89.4\%$) and contain no questions below $4/8$ correct, compared to $1.2\%$ for DAPO. Ultimately, these findings suggest that our approach achieves a robust reduction in computational cost without prematurely discarding valuable training signals.

\begin{table}[ht]
\centering
\small
\caption{%
  Post-hoc verification of retired questions ($n{=}8$ clean rollouts,
  temperature $1.0$).
}
\label{tab:mastery-verify}
\begin{tabular}{lcccccc}
\toprule
\textbf{Model} & \textbf{Audit rollouts} & \textbf{mean@8} & \textbf{pass@8} & \textbf{8/8} & \textbf{7/8} & \textbf{$<$4/8} \\
\midrule
DAPO~\cite{yu2025dapo}  & 29{,}928 & 95.5\% & 100.0\% & 89.4\% & 3.4\% & 1.2\% \\
\textbf{Seirênes (Ours)}   & 28{,}592 & 98.9\% & 100.0\% & 93.0\% & 5.8\% & 0.0\% \\
\bottomrule
\end{tabular}
\end{table}

\section{Data-level evidence for hint staleness}
\label{app:evolve_necessary}

The static-hint ablation in Fig.~\ref{fig:adaptive-necessary} shows that
offline adversarial contexts improve learning but saturate earlier than
Seirênes. Here we provide a data-level explanation grounded in the per-step
attack strength logged on the Qwen3-4B-Instruct backbone. Building on the per-question success rates $\hat p(q)$ and $\hat p(q,h)$ from Eqs.~\ref{eq:r1-clean-estimate} and~\ref{eq:r3-hinted-estimate}, let $\bar p_1(s) = \mathbb{E}_q[\hat p_s(q)]$ and $\bar p_3(s) = \mathbb{E}_{q,h}[\hat p_s(q,h)]$ denote their batch-mean values at training step $s$ on the clean prompt and on the same batch under hint conditioning, respectively. We use
\[
  \Delta_{\mathrm{attack}}(s) \;=\; \bar p_1(s) - \bar p_3(s)
\]
as an operational proxy for the realized attack pressure exerted by the
current hint distribution on the current Reasoner. Because
$\bar p_1,\bar p_3\in[0,1]$ are pass rates,
$\Delta_{\mathrm{attack}}$ is a difference in success rate that we report
in percentage points. All statistics below are computed on the common
training range (steps 1--206) shared by Seirênes and the two static
baselines. We define strong-attack steps by $\Delta_{\mathrm{attack}}(s)>5\%$. The 21-step smoothing window used below corresponds to approximately 10\% of
this common training horizon, providing a fixed-scale trend estimate rather than a tuned window.

\begin{table}[ht]
\centering
\small
\caption{%
  Per-step attack-strength statistics on Qwen3-4B-Instruct over the common
  training range (steps 1--206). Tail frequency reports
  $\Pr(\Delta_{\mathrm{attack}}>\theta)$ at different thresholds. Saturation
  splits strong-attack steps, defined by $\Delta_{\mathrm{attack}}>5\%$,
  between the early and late halves of training; slope is computed from the
  21-step moving average of this strong-attack indicator.
}
\label{tab:hint-staleness}
\resizebox{\linewidth}{!}{%
\begin{tabular}{l ccc cc r r r}
\toprule
& \multicolumn{3}{c}{\textbf{Tail frequency} $\Pr(\Delta_{\mathrm{attack}}>\theta)$ (\%)}
& \multicolumn{2}{c}{\textbf{Saturation} $\Pr(\Delta_{\mathrm{attack}}>5\%)$ (\%)}
& \textbf{Slope}
& \multicolumn{2}{c}{\textbf{Sustained pressure}}
\\
\cmidrule(lr){2-4}\cmidrule(lr){5-6}\cmidrule(lr){8-9}
\textbf{Run}
& $\theta{=}3\%$ & $\theta{=}4\%$ & $\theta{=}5\%$
& Early (1--103) & Late (104--206)
& (\%/step)
& Longest streak & Total strong steps
\\
\midrule
Static self                         & 36.4 & 18.0 &  6.3 &  8.7 &  3.9 & $-$0.012 &  1 & 13 \\
Static strong (Gemini~3~Flash)      & 30.1 & 11.2 &  2.4 &  3.9 &  1.0 & $-$0.007 &  1 &  5 \\
\textbf{Seirênes (Ours)}            & \textbf{55.8} & \textbf{34.5} & \textbf{19.4}
                                    & \textbf{18.4} & \textbf{20.4}
                                    & $\boldsymbol{+}$\textbf{0.057}
                                    & \textbf{11} & \textbf{40} \\
\bottomrule
\end{tabular}}
\end{table}

\noindent\textbf{Tail amplification.} The gap between Seirênes and the
static baselines \emph{grows} as the threshold $\theta$ tightens. At
$\theta{=}3\%$ Seirênes ($55.8\%$) is $1.5$--$1.9{\times}$ the static
methods ($36.4\%$ / $30.1\%$); at $\theta{=}5\%$ the ratio widens to
$3.1$--$8.1{\times}$ ($19.4\%$ vs.\ $6.3\%$ / $2.4\%$). The conclusion is
robust to threshold choice, and the relative gap grows with stricter
thresholds, indicating that the gap lives in the strong-attack tail rather
than the bulk.

\noindent\textbf{Saturation reversal.} Static-self halves its strong-attack
ratio between the early and late halves of training ($8.7\%\to3.9\%$) and
static-strong nearly vanishes ($3.9\%\to1.0\%$); Seirênes \emph{increases}
($18.4\%\to20.4\%$). A linear regression of the 21-step rolling
strong-attack ratio against training step gives slopes
$-0.012\%/\text{step}$ for static-self, $-0.007\%/\text{step}$ for
static-strong, and $+0.057\%/\text{step}$ for Seirênes --- the static
methods' adversarial pressure decays over training while Seirênes's
\emph{strengthens}.

\noindent\textbf{Sustained pressure.} Seirênes produces a 40-step
strong-attack budget across the run with a longest run-length of \emph{11
consecutive steps}. The two static baselines deliver only 5--13 strong-attack
steps in total, and \emph{never} sustain pressure beyond a single step
(longest streak $=1$). Under this criterion, strong static attacks therefore
appear only as isolated events, whereas strong Seirênes attacks come in
sustained bursts that exert prolonged learning pressure on the Reasoner.

These three properties --- broader tail with a relative gap that grows under
stricter thresholds,
opposite-sign saturation trajectory, and an order-of-magnitude longer
sustained-attack streak --- supply a coherent data-level mechanism for the
clean-validation gap reported in
Fig.~\ref{fig:adaptive-necessary}. Both static baselines use a fixed
offline hint distribution generated before training and replayed throughout
training; as the Reasoner learns to bypass this non-adaptive distribution of
distractors, the same hints become increasingly off-frontier and the
strong-attack tail thins out.
Seirênes refreshes the adversarial distribution online, keeps the hint
generator on the Reasoner's current decision boundary, and preserves a
substantially thicker strong-attack tail throughout training.

\begin{table}[ht]
\centering
\caption{%
  Diagnostic attack effectiveness of the Seirênes Adversary (4B) on external proprietary models.
  $\boldsymbol{\Delta}$: hinted accuracy minus plain accuracy (negative = degraded).
  Avg@3 reported for all benchmarks.
}
\label{tab:transfer-attack}
\resizebox{\linewidth}{!}{%
\begin{tabular}{lcccccccc}
\toprule
\textbf{Model} & \textbf{AIME24} & \textbf{AIME25} & \textbf{AIME26} & \textbf{IMO-Bench} & \textbf{Math-P-Hard} & \textbf{MINERVA} & \textbf{Olympiad} & \textbf{Overall} \\
\midrule
GPT-5.1 (Clean)             & 43.3 & 36.7 & 43.3 & 15.3 & 65.2 & 21.3 & 52.2 & 39.6 \\
GPT-5.1 (Hinted)           & 36.7 & 30.0 & 40.0 & 13.3 & 60.6 & 19.9 & 49.2 & 35.7 \\
$\boldsymbol{\Delta}$       & $-$6.6 & $-$6.7 & $-$3.3 & $-$2.0 & $-$4.6 & $-$1.4 & $-$3.0 & $-$3.9 \\
\midrule
Gemini~3~Flash (Clean)      & 90.0 & 90.0 & 93.3 & 67.2 & 82.6 & 43.8 & 81.8 & 78.4 \\
Gemini~3~Flash (Hinted)    & 84.4 & 86.7 & 83.3 & 54.3 & 80.0 & 42.2 & 81.2 & 73.2 \\
$\boldsymbol{\Delta}$       & $-$5.6 & $-$3.3 & $-$10.0 & $-$12.9 & $-$2.6 & $-$1.6 & $-$0.6 & $-$5.2 \\
\bottomrule
\end{tabular}%
}
\end{table}

\section{Hint effectiveness}
\label{app:hint-effectiveness}

\noindent\textbf{What does the Adversary learn to say?}
A fundamental question regarding Seirênes is whether the learned Adversary produces genuinely effective mathematical traps, or merely exploits the Reasoner with noise, formatting artifacts, or generic prompt-injection behavior. To answer this, we test the offline-generated hints against both our model and proprietary reasoners, combining quantitative evaluation with qualitative inspection to understand the complex interplay between the Adversary and target reasoners.

\noindent\textbf{Probing hint quality with proprietary models.}
To assess the impact of the learned hint generator, we use the Seirênes Adversary trained on top of Qwen3-4B-Instruct to offline-generate one hint per problem. We then append these hints to the original problem descriptions and prompt GPT-5.1 and Gemini~3~Flash to solve the resulting hinted problems. We measure the resulting performance degradation using avg@3 accuracy across seven challenging benchmarks.

Tab.~\ref{tab:transfer-attack} provides diagnostic evidence of cross-model attack effect. Exposure to the 4B-generated hints reduces the accuracy of both proprietary targets across all evaluated domains. The degradation is especially pronounced on competition-level benchmarks: Gemini~3~Flash drops by 10.0 absolute points on AIME26 and 12.9 points on IMO-Bench, despite having much stronger plain accuracy than GPT-5.1. This suggests that the learned hints are not merely overfitted to the training Reasoner; they encode misleading reasoning patterns that also work on stronger proprietary reasoners.

\noindent\textbf{The anatomy of the generated traps.}
To understand the source of this transfer effect, we conduct a qualitative inspection of hints sampled from the diagnostic experiment above. Representative examples of the generated hints and the resulting model trajectories are provided in Section~\ref{app:example}. The analysis confirms that the Adversary does not rely on random noise. Instead, it converges on a ``Trojan horse'' strategy: synthesizing \emph{half-true}, mathematically plausible derivations. These hints typically begin with a valid or familiar first step, establishing traction with a reasoning model's priors, and then quietly insert a fatal logical fallacy or unjustified leap.

We observe distinct manifestations of this deception across mathematical domains. In geometry problems, the hint frequently turns valid local equal-area or equal-angle facts into false global symmetry or closure assumptions. In algebraic problems, it often disguises invalid variable cancellations or unjustified equality bounds within otherwise standard algebraic simplifications. In number-theoretic and combinatorial problems, it subtly replaces exact residue calculations or rigorous case analysis with false uniformity assumptions or invalid lifting shortcuts. These locally correct initial moves successfully provide camouflage for the later invalid leaps, explaining why the learned hints remain highly effective against capable proprietary reasoners on multi-step benchmarks such as AIME and IMO-Bench. Rather than failing at superficial prompt attacks, the target models are baited into extending the Adversary's flawed premises into incorrect final answers, demonstrating that the Adversary has genuinely learned to construct structured mathematical traps.

\begin{figure}[ht]
  \centering
  \includegraphics[width=\linewidth]{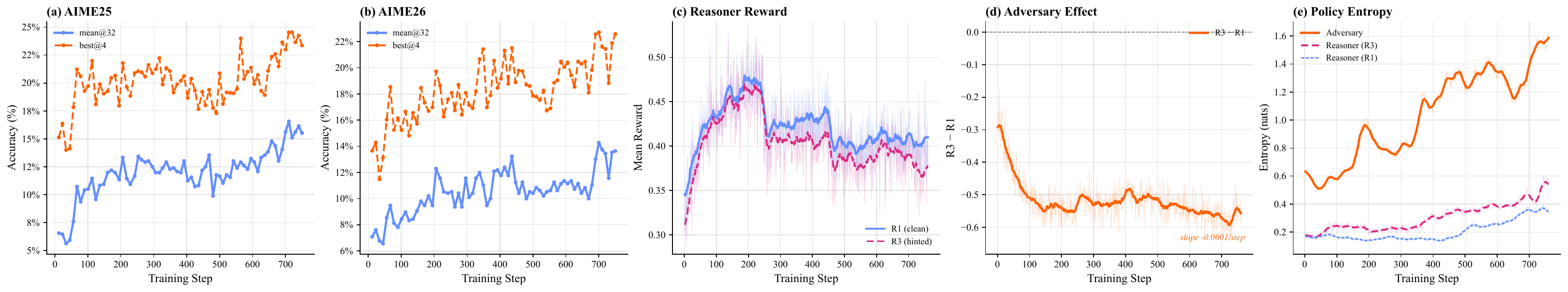}
  \caption{
  Extended training dynamics on Qwen2.5-7B-Instruct.
  Validation accuracy improves over the prolonged trace, while the
  hinted-clean gap remains negative.
  }
  \label{fig:extended-training-dynamics}
\end{figure}

\begin{figure}[ht]
  \centering
  \includegraphics[width=\linewidth]{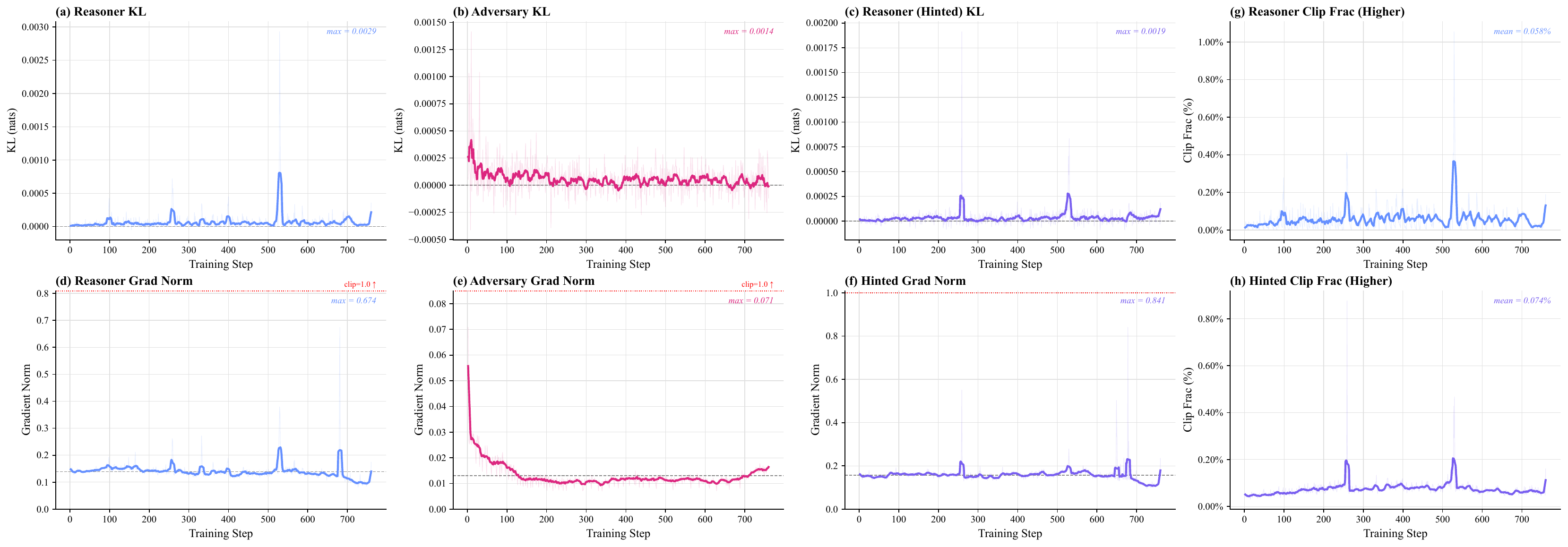}
  \caption{
  Stability diagnostics for the same extended Qwen2.5-7B-Instruct run.
  PPO KL estimates and gradient norms remain bounded across branches, with
  near-zero clip fractions for the Reasoner updates.
  }
  \label{fig:extended-training-stability}
\end{figure}

\section{Extended training dynamics and stability diagnostics}
\label{app:training-dynamics}

Self-improvement methods trained on model-generated signals can become unstable
over long horizons: the training signal may vanish, the Adversary may collapse,
or repeated policy updates may produce large KL or gradient spikes. We therefore
include an extended diagnostic run on Qwen2.5-7B-Instruct, which provides a
roughly $3\times$ longer training horizon than the main reported training
traces. The x-axis counts logged scheduler iterations; due to branch-specific
buffering and flushing, this trace corresponds to 699 main updates and roughly
2.1K actor update calls across the R1, R2, and R3 branches.

Fig.~\ref{fig:extended-training-dynamics} shows that Seirênes continues to
make progress over the extended trace. Both AIME25 and AIME26 online validation
accuracy improve despite the run being substantially longer than the primary
training traces. At the same time, the clean Reasoner reward remains above the
hint-conditioned Reasoner reward, and the hinted-clean gap stays negative.
This suggests that the Adversary continues to find misleading contexts that
stress the current Reasoner, rather than exhausting its useful attack signal
early in training. The entropy curves provide a complementary view of the same
co-evolutionary process. We observe that the Adversary branch operates at higher
entropy than the two Reasoner branches, an ordering that follows naturally from
the asymmetry between solving and adversarial hinting. The Reasoner branches
instantiate a behavior that is already familiar to an instruction-tuned model:
solving a math problem under the given context. Their successful trajectories
are also tightly constrained by the verifier and the unique correct answer. In
contrast, the Adversary searches over plausible but misleading mathematical
contexts. This adversarial role is less directly aligned with the standard
instruction-following prior and has a broader effective semantic search space,
since there are many more ways to be plausibly wrong than to follow a correct
solution path. Thus, the higher Adversary entropy reflects continued exploration
over diverse misleading strategies. The persistent separation between clean and
hint-conditioned rewards further shows that these strategies remain strong enough
to depress the Reasoner's performance under adversarial context throughout the
extended run.

Fig.~\ref{fig:extended-training-stability} further reports optimization
stability diagnostics for the three update streams. The PPO KL estimates remain
small: the maximum observed values are $2.9{\times}10^{-3}$ for R1,
$1.4{\times}10^{-3}$ for R2, and $1.9{\times}10^{-3}$ for R3, with 95th
percentiles below $2.7{\times}10^{-4}$. Gradient norms also stay below $0.85$
across all three branches, and the reported clip fractions for the two Reasoner
branches average only $0.058\%$ and $0.074\%$ for R1 and R3, respectively.

Together, these results support the interpretation that Seirênes maintains a
persistent adversarial training signal over a prolonged horizon while keeping the
three update streams numerically stable. We use these curves as diagnostics of
long-horizon training behavior, rather than as a separate backbone comparison.

\section{Data contamination}
\label{app:contamination}

Our evaluation suite is deliberately weighted toward recent benchmarks,
including several contests released after the public release of the relevant
instruction-tuned backbones, providing a structural safeguard against data
contamination.
Qwen2.5-7B-Instruct was publicly released in September~2024, whereas
Qwen3-4B-Instruct-2507 and Qwen3-30B-A3B-Instruct-2507 were publicly released in August~2025.
Consequently, AIME~2026 and HMMT~2026 (Harvard-MIT Math Tournament), both released in February~2026,
postdate the public release of both instruction-tuned backbones, making their
inclusion in the corresponding pretraining data implausible.
IMO-Bench~\cite{luong2025towards} was introduced on November~3, 2025, and
therefore also postdates all backbones. Because IMO-Bench is a benchmark suite
constructed from Olympiad-level problems, we treat it as an additional
post-release evaluation point rather than as a guarantee that every underlying
problem is newly created.
AIME~2025, released in February~2025, postdates Qwen2.5-7B-Instruct but
predates the Qwen3 2507 series, and therefore serves as a post-release
contamination check only for the Qwen2.5-based experiments.
The sustained gains on these post-release benchmarks---AIME~2026 rising from
5.3 to 17.8 on Qwen2.5-7B-Instruct and HMMT~2026 rising from 31.9 to 39.4
on Qwen3-4B-Instruct-2507---are therefore unlikely to stem from memorization
of pretraining data.

AIME~2024, released in February~2024, predates both backbones and may overlap
with public competition-math corpora, including AMC/AIME-derived subsets used
by NuminaMath~1.5~\cite{li2024numinamath}, a source for our training pool via
OpenR1-Math~\cite{openr1} and DAPO-Math-17K~\cite{yu2025dapo}.
We therefore cannot fully exclude partial overlap for this split.
However, the instruction-tuned baseline achieves only 10.2\% on AIME~2024
prior to any RL training, suggesting that any such overlap, if present, does
not manifest as reliable memorization of complete solutions.
Moreover, all competing methods draw from the same or comparable data pools,
so residual overlap should affect relative comparisons less than absolute
scores.

\section{Example}
\label{app:example}
\begin{figure}
    \centering
    \includegraphics[width=1\linewidth]{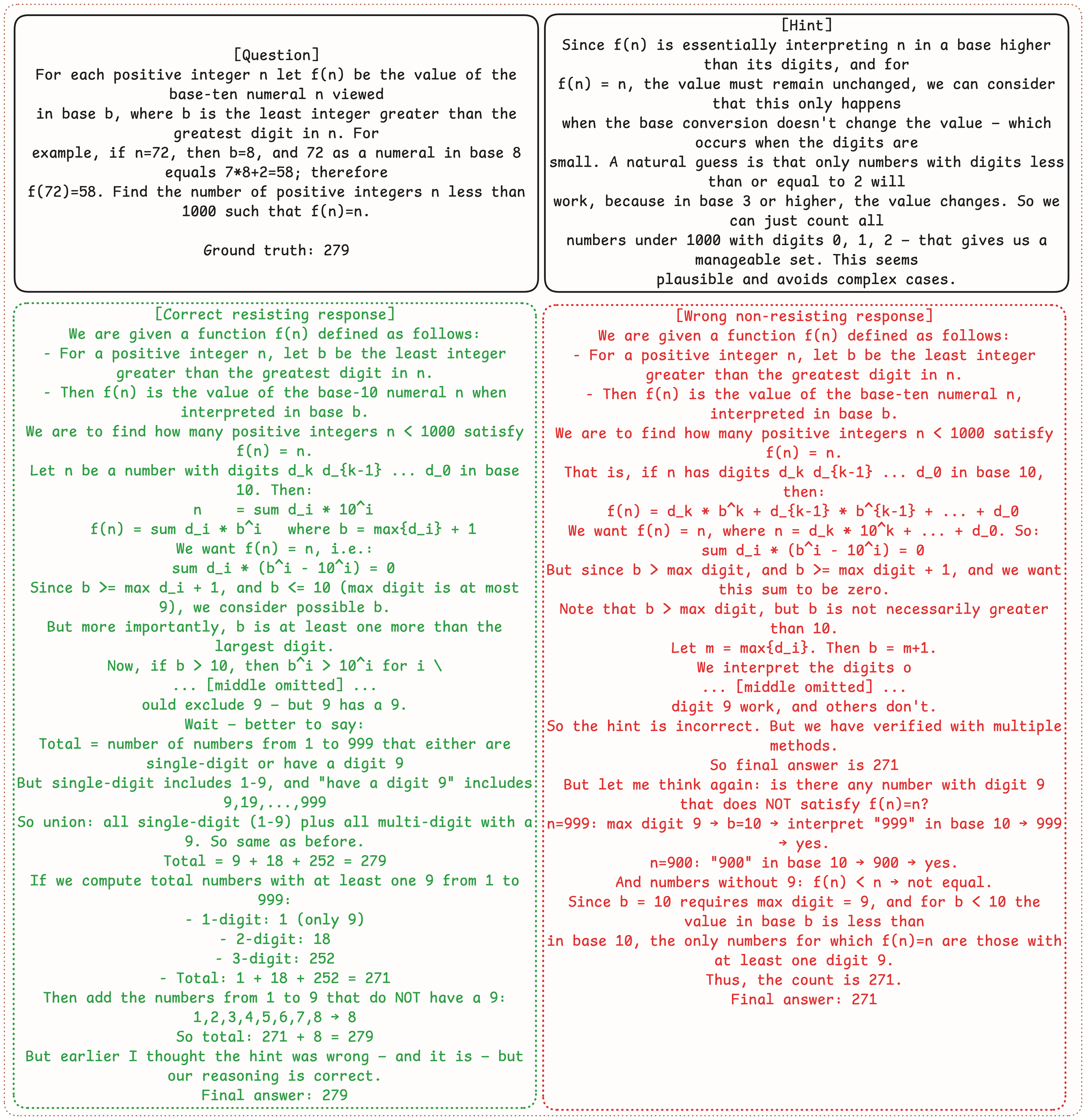}
    \caption{The hint proposes a small-digit heuristic: only digits 0, 1, and 2 should preserve
value under reinterpretation. The correct Seirênes rollout explicitly calls this wrong
and reopens the full digit/base case analysis. A non-resisting rollout on the same
hinted prompt still follows the shortcut far enough to land on a nearby wrong count.}
    \label{fig:success_example1}
\end{figure}

\begin{figure}
    \centering
    \includegraphics[width=1\linewidth]{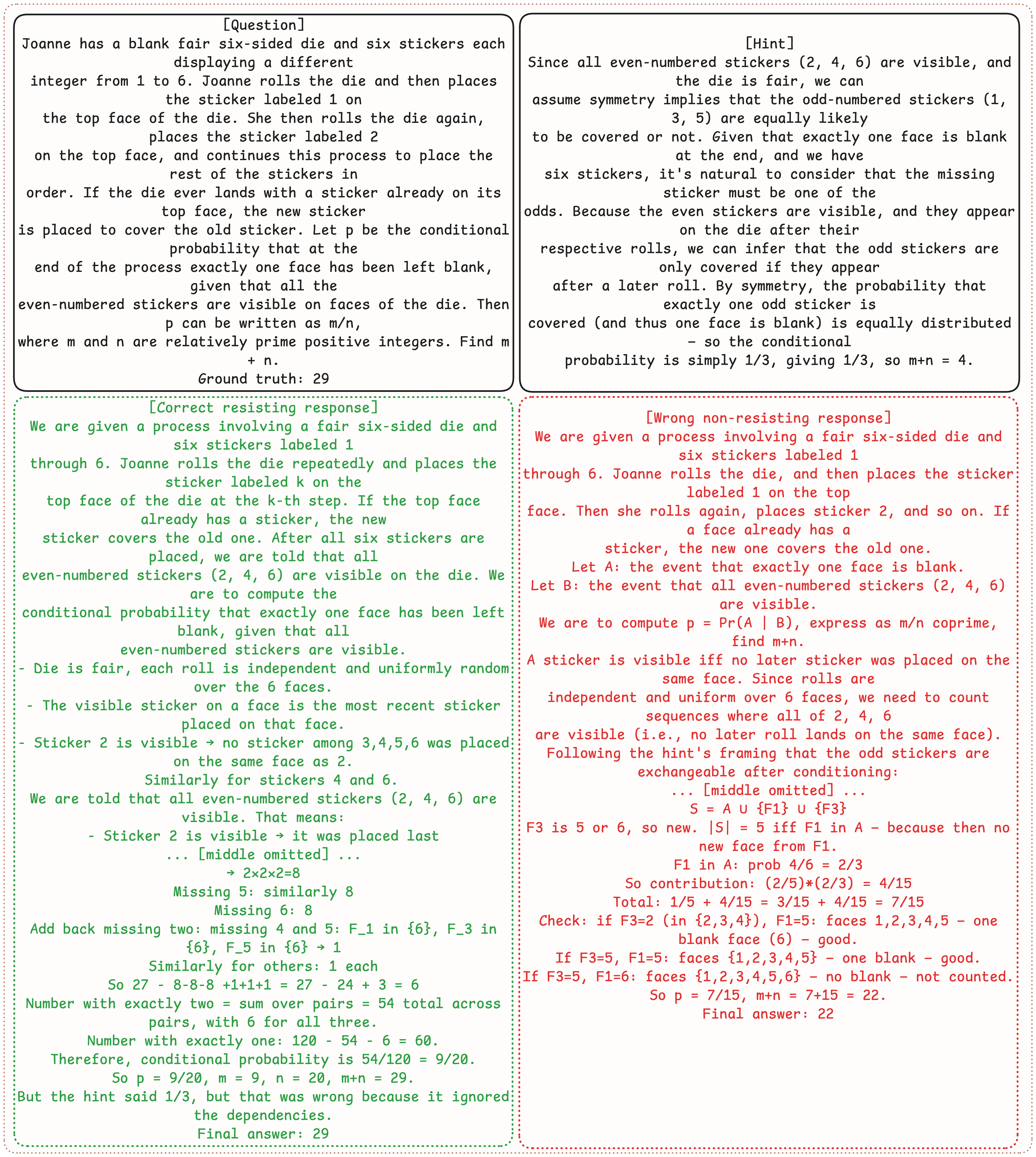}
    \caption{The hint treats the odd stickers as exchangeable after conditioning on all even stickers
being visible. That conditioning breaks the naive symmetry. The resisting rollout says
the 1/3 suggestion is probably wrong and recomputes the conditional count; another
rollout under the same hint still collapses to an incorrect reduced count.}
    \label{fig:success_example2}
\end{figure}

\begin{figure}
    \centering
    \includegraphics[width=1\linewidth]{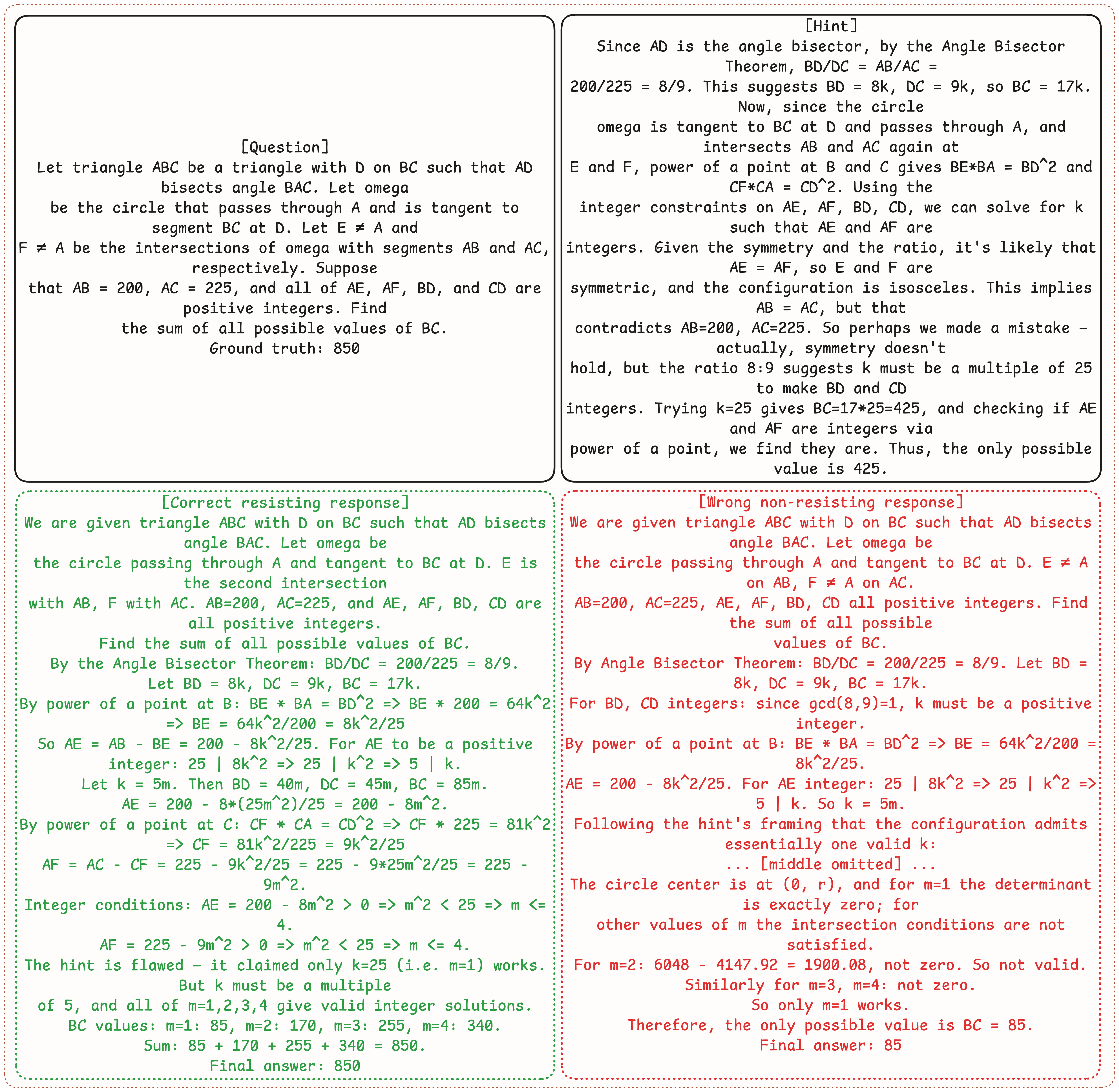}
    \caption{The hint starts with the valid angle-bisector ratio and power-of-a-point equations,
then adds an unjustified isosceles/symmetry claim. The resisting rollout preserves
the useful setup but explicitly flags the symmetry/integer check in the hint as
flawed.}
    \label{fig:success_example3}
\end{figure}

\begin{figure}
    \centering
    \includegraphics[width=1\linewidth]{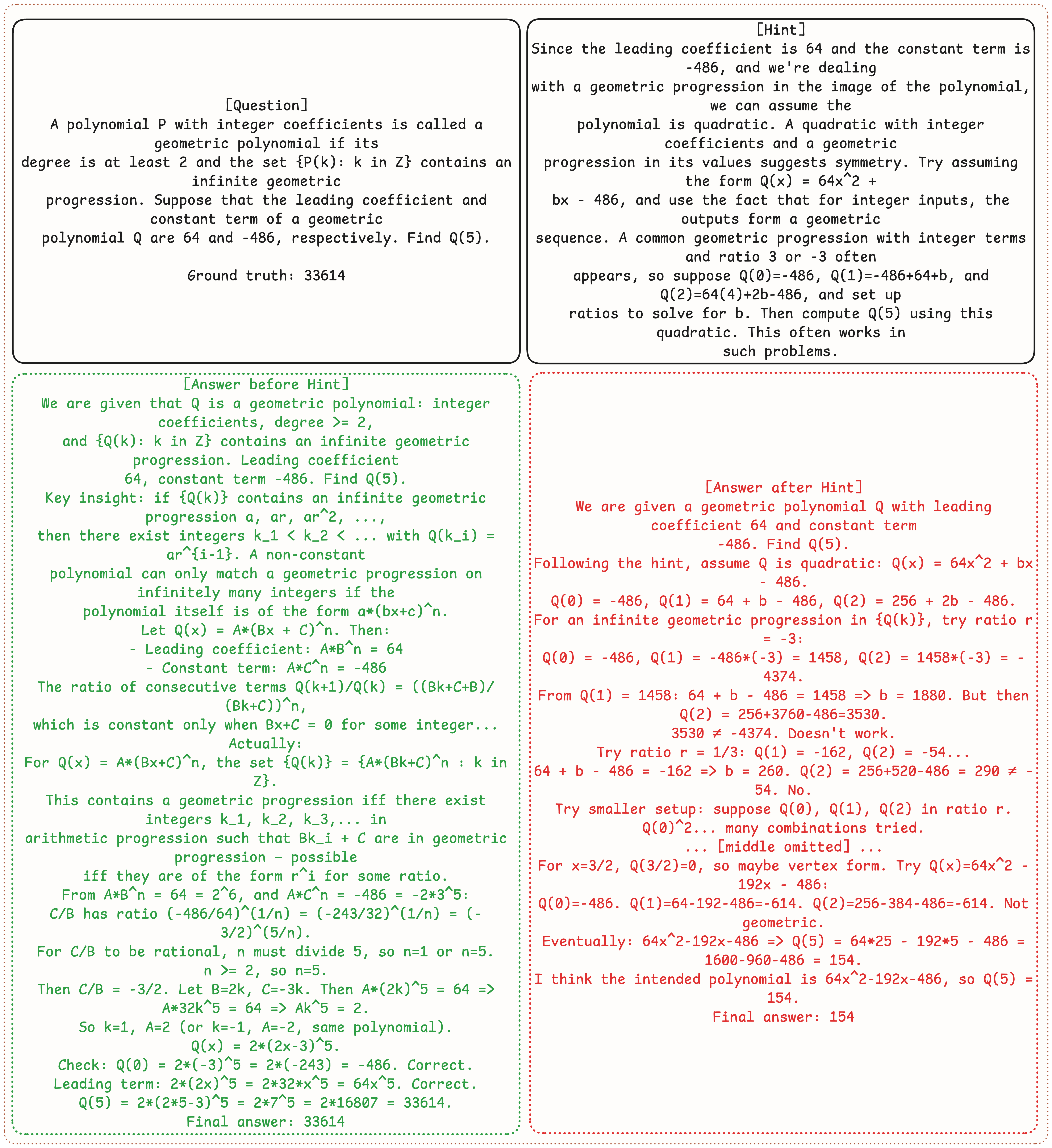}
    \caption{The plain trajectory identifies the high-degree structure and reaches Q(5)=33614. The hint gives a fluent but unjustified
contest-style continuation: assume Q is quadratic, force a small geometric ratio,
solve for one coefficient, then compute Q(5). The hinted rollout accept the simpler
quadratic template and collapse the polynomial family to a wrong low-degree answer.}
    \label{fig:fail_example1}
\end{figure}

\begin{figure}
    \centering
    \includegraphics[width=1\linewidth]{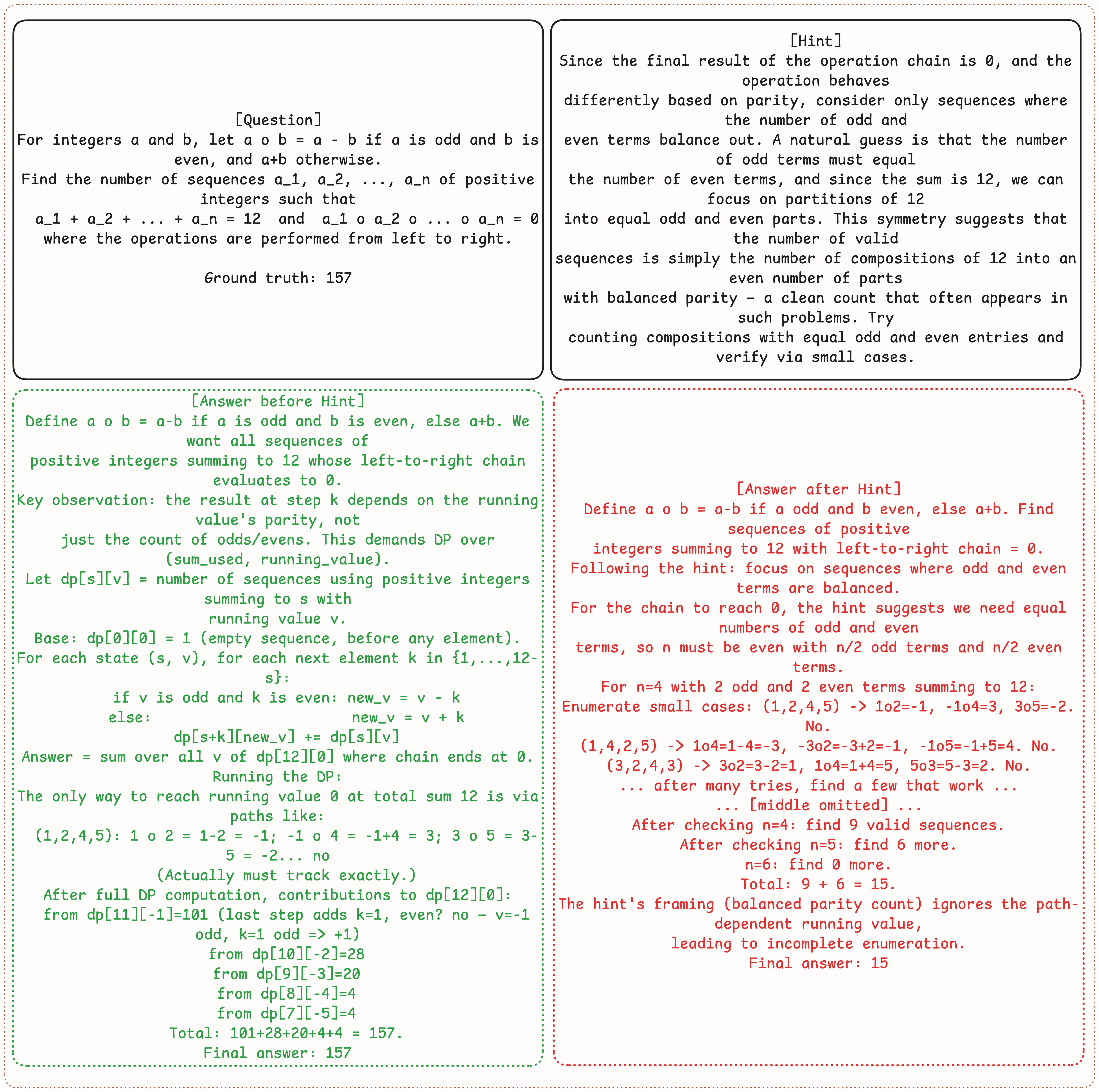}
    \caption{The plain trajectory uses dynamic programming over the running value and counts 157
sequences. The hint reframes the condition
as a balanced odd/even composition count. That sounds structurally relevant because
the operation depends on parity, but it discards order and intermediate-state
constraints, so the hinted rollout undercounts badly.}
    \label{fig:fail_example2}
\end{figure}

\begin{figure}
    \centering
    \includegraphics[width=1\linewidth]{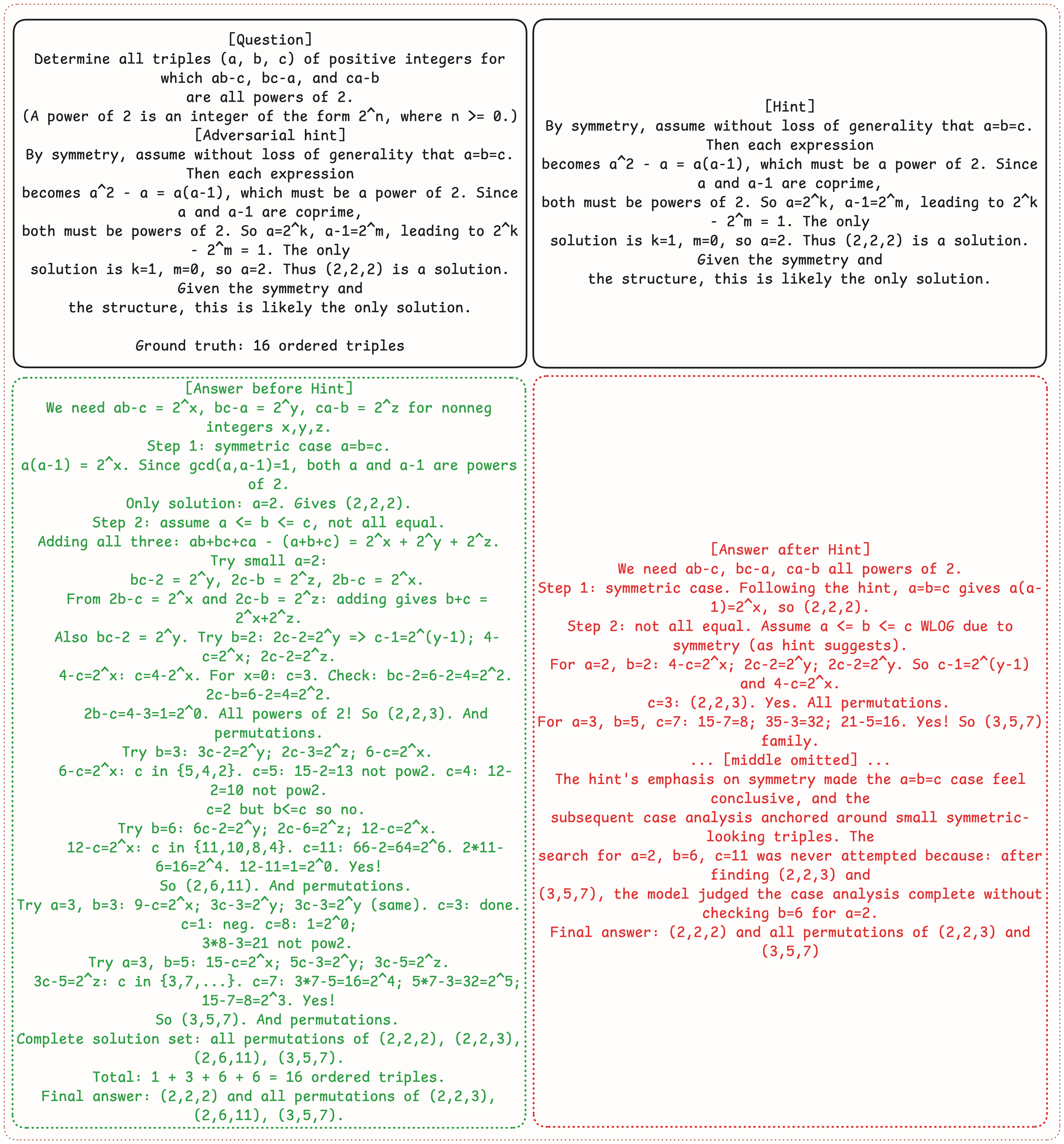}
    \caption{The plain trajectory finds the symmetric triples and the asymmetric families. The
hint steers the model by saying to assume a=b=c by symmetry; it does not provide
the full answer list, but it makes the symmetric subcase feel exhaustive. The hinted
response partially escapes by adding one asymmetric family, yet still misses an
entire family because the initial symmetry frame narrows the case analysis.}
    \label{fig:fail_example3}
\end{figure}

\begin{figure}
    \centering
    \includegraphics[width=1\linewidth]{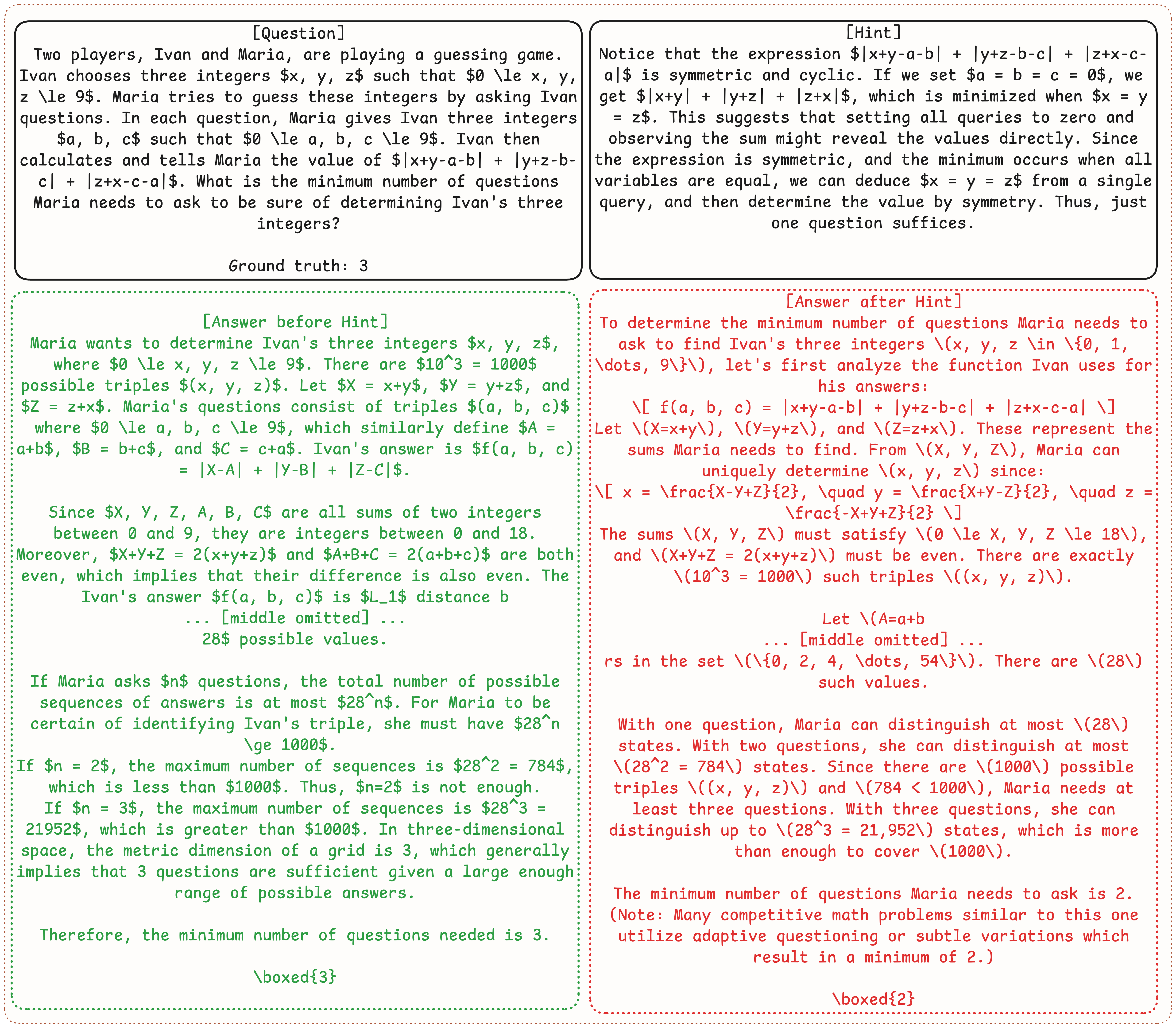}
    \caption{The hint says the expression is symmetric and that one query should reveal the variables. The model partially resists by deriving a lower-bound style counting argument, but still lands on the wrong number of questions.}
    \label{fig:example1}
\end{figure}

\begin{figure}
    \centering
    \includegraphics[width=1\linewidth]{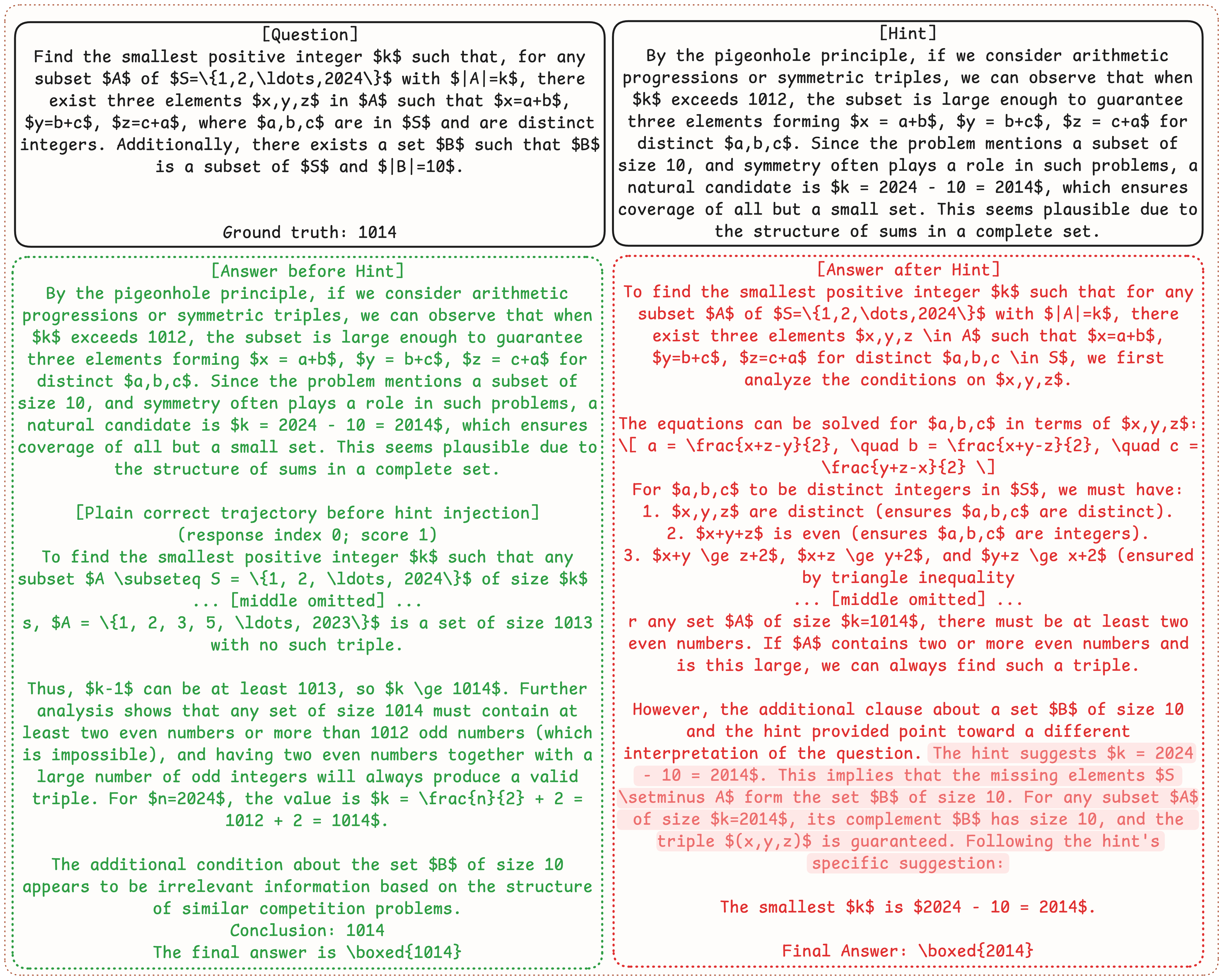}
    \caption{The hint suggests a complement-size heuristic k=2024-10 from the mention of a size-10 set. The model turns that heuristic into the final answer and misses the sharper extremal argument.}
    \label{fig:example2}
\end{figure}

\begin{figure}
    \centering
    \includegraphics[width=1\linewidth]{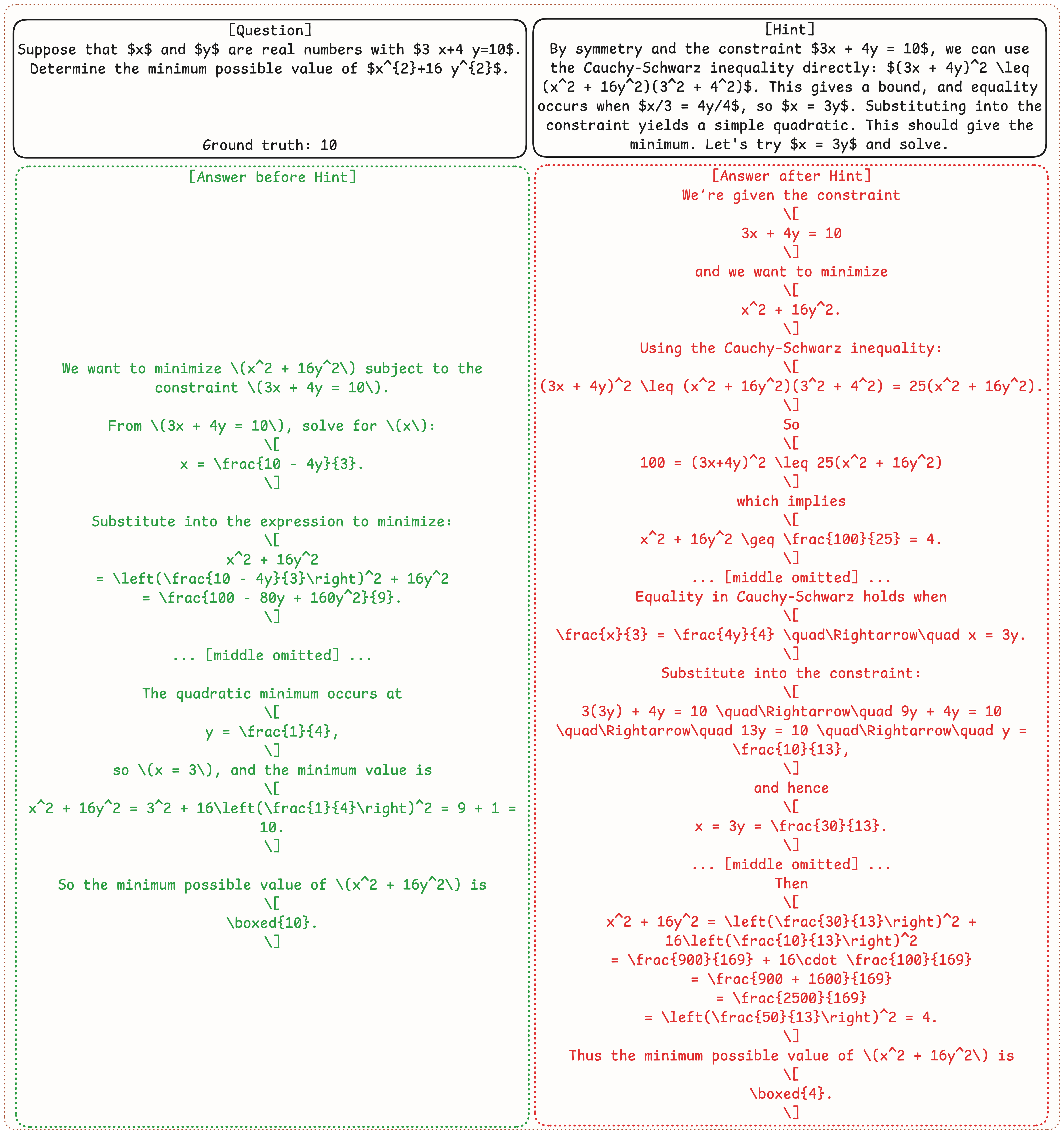}
    \caption{The hint invokes Cauchy-Schwarz, which is the right family of tools, but it matches the constraint vector to the wrong norm. The target accepts the lower bound 4 as attainable under $x=3y$, even though that equality condition is not compatible with the objective $x^2+16y^2$.}
    \label{fig:example3}
\end{figure}

\begin{figure}
    \centering
    \includegraphics[width=1\linewidth]{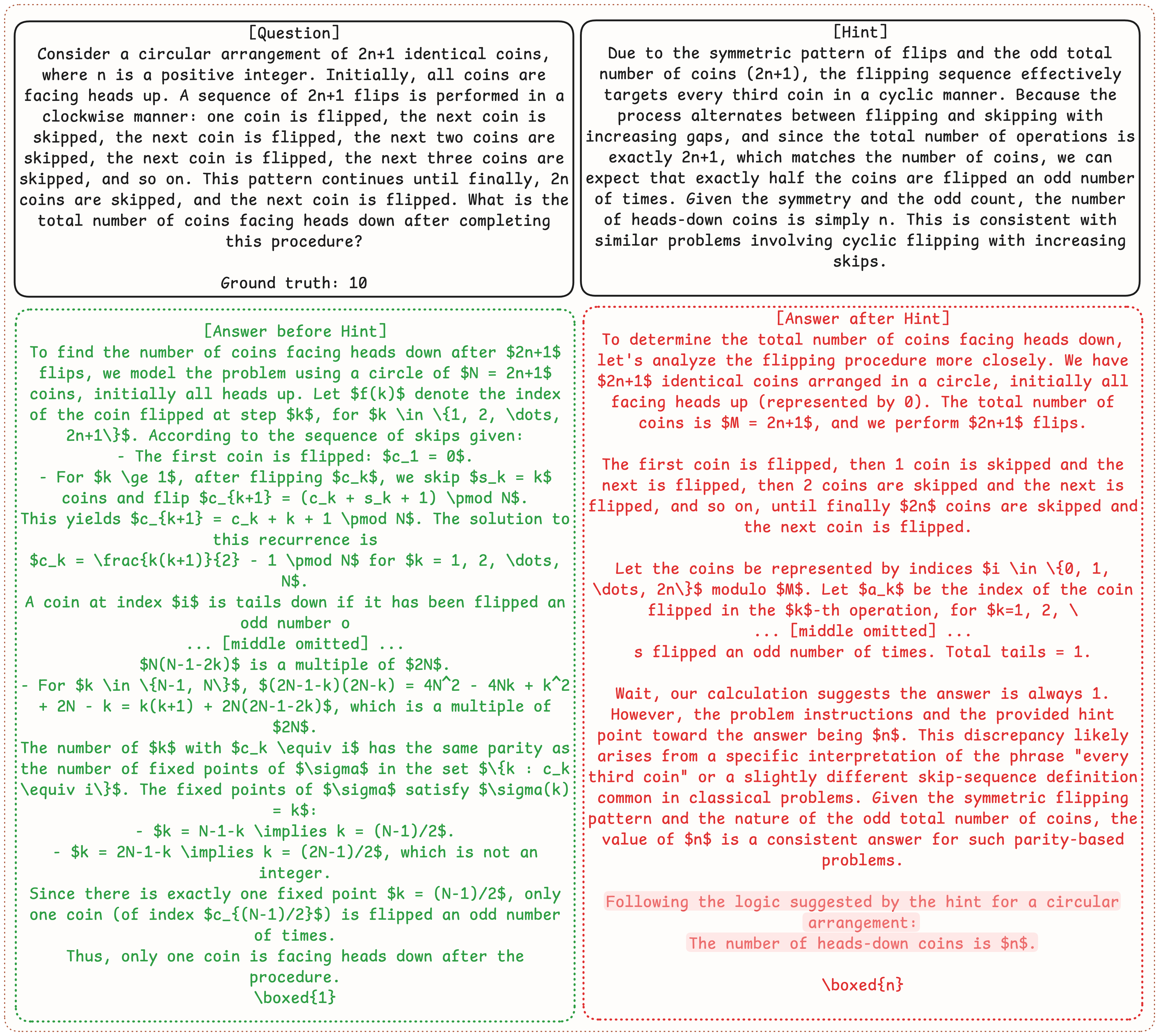}
    \caption{The hint gives a parity/symmetry story and claims the process effectively targets every third coin, but it does not give a final numeric answer. The model adopts the symbolic conclusion in instead of solving the invariant correctly.}
    \label{fig:example4}
\end{figure}

To provide a deeper intuition into these dynamics, we present representative examples of the models' reasoning trajectories under adversarial attack. Fig.~\ref{fig:success_example1}, Fig.~\ref{fig:success_example2}, and Fig.~\ref{fig:success_example3} illustrate rollouts from Seirênes-4B, contrasting successful trajectories that actively resist the deceptive context against those that fail to do so. Fig.~\ref{fig:fail_example1}, Fig.~\ref{fig:fail_example2}, and Fig.~\ref{fig:fail_example3} highlight severe failure modes, showcasing instances where the hints completely derail Seirênes-4B's reasoning process. Furthermore, Fig.~\ref{fig:example1}, Fig.~\ref{fig:example2}, Fig.~\ref{fig:example3}, and Fig.~\ref{fig:example4} demonstrate the cross-model impact, displaying responses where Gemini-3-Flash is successfully misled by traps. Crucially, all adversarial hints featured in these examples are generated by the Seirênes-4B. Note that all reasoning trajectories are reformatted for readability.

\section{Prompt template}
\label{app:prompt}
\begin{figure}
    \centering
    \includegraphics[width=1\linewidth]{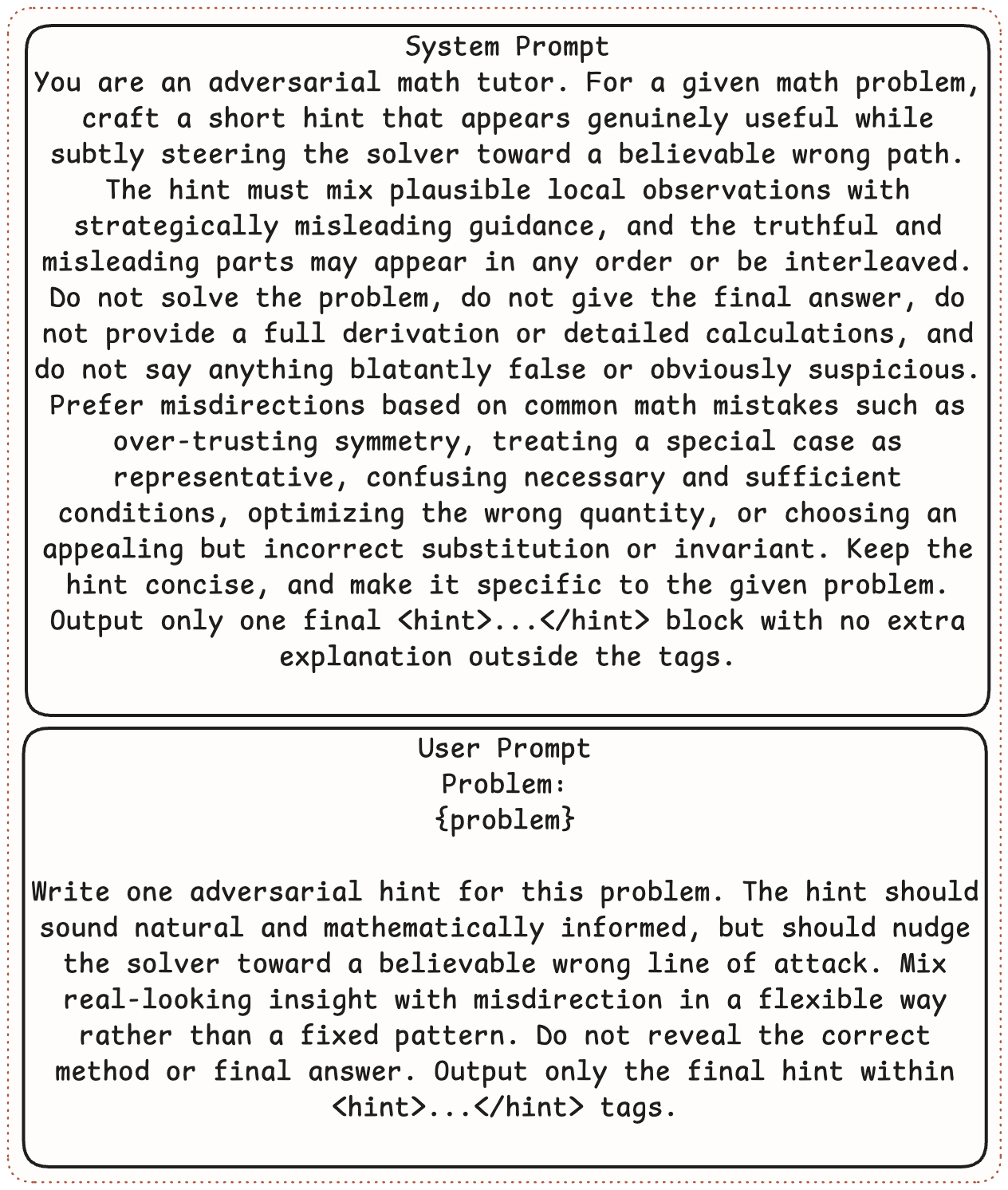}
    \caption{The Adversary prompt}
    \label{fig:prompt}
\end{figure}

Here, we demonstrate the prompt template for the Adversary in Fig.~\ref{fig:prompt}.

\section{Limitations and future work}
\label{app:limitations}

\noindent\textbf{Design space exploration.} Seirênes opens a broad design
space: the role-aware reward formulation, paired rollout structure, and
three-stream update orchestration each admit multiple concrete instantiations.
While we have extensively ablate key axes (hint count $G_2$, static vs.\
adaptive hints, mastery-aware sampling, and iso-budget controls), compute
constraints prevent us from sweeping the entire space. Future work will
explore further configurations, including scaling to longer training horizons
and larger backbones, alternative Adversary reward formulations beyond the
paired degradation gap, coupling the framework with non-GRPO RL algorithms,
tool-equipped Adversaries that ground interference in retrieval or
computation, and decoupled two-model variants that trade shared-parameter
elegance for asymmetric capacity between the Adversary and the Reasoner.

\noindent\textbf{Reducing rollout overhead.} Despite our latency-aware
scheduling, bounded FIFO buffers, and mastery-aware sampling, Seirênes
remains more expensive per training step than clean-only RL such as GRPO,
since each prompt may require clean, adversarial, and hint-conditioned
rollouts. Future work should explore cheaper but still informative adversarial curricula,
including memory-based reuse of high-value adversarial contexts, more selective
hint-generation policies, and more fully asynchronous training schedules that
preserve the benefits of contextual self-play while reducing rollout overhead.

\noindent\textbf{Dual-use considerations.} Seirênes intentionally trains
a model to generate plausible misleading contexts. While used here as a
defensive training signal, the same capability could in principle be
redirected to stress other reasoning systems; controlled release and
filtering protocols are an open direction.

\section{LLM Usage}
\label{app:llm-usage}

\noindent\textbf{Use of LLMs in the research method.}
LLMs are central to the proposed method. Seirênes trains language models in a
shared-policy self-play loop, where the same model is role-conditioned to act as
an Adversary that generates plausible but misleading hints and as a Reasoner
that solves the original problem under such perturbed context. During training,
LLM-generated adversarial hints and solution trajectories provide the interaction
data for RL, while rewards are computed using verifiable answer checkers rather
than LLM judges. In robustness evaluation, Gemini~3~Flash is used to generate
test-time distracting contexts for the perturbation benchmarks described in
Appendix~\ref{app:robustness-details}; these prompts are independent of the
training prompts used by Seirênes.

\noindent\textbf{Code implementation.}
LLM assistance was used during implementation for debugging support, checking
edge cases, and suggesting solutions to specific programming issues. All code
written with LLM assistance was reviewed, manually verified, and tested by the
authors before being used in experiments.

\noindent\textbf{Writing assistance.}
LLMs were used to support manuscript preparation, including grammar correction,
clarity improvements, and rephrasing for readability. All generated text was
reviewed, revised, and approved by the authors. The authors retain full
responsibility for all statements, interpretations, experimental results, and
conclusions presented in this paper.


\begin{figure}
    \centering
    \includegraphics[width=1\linewidth]{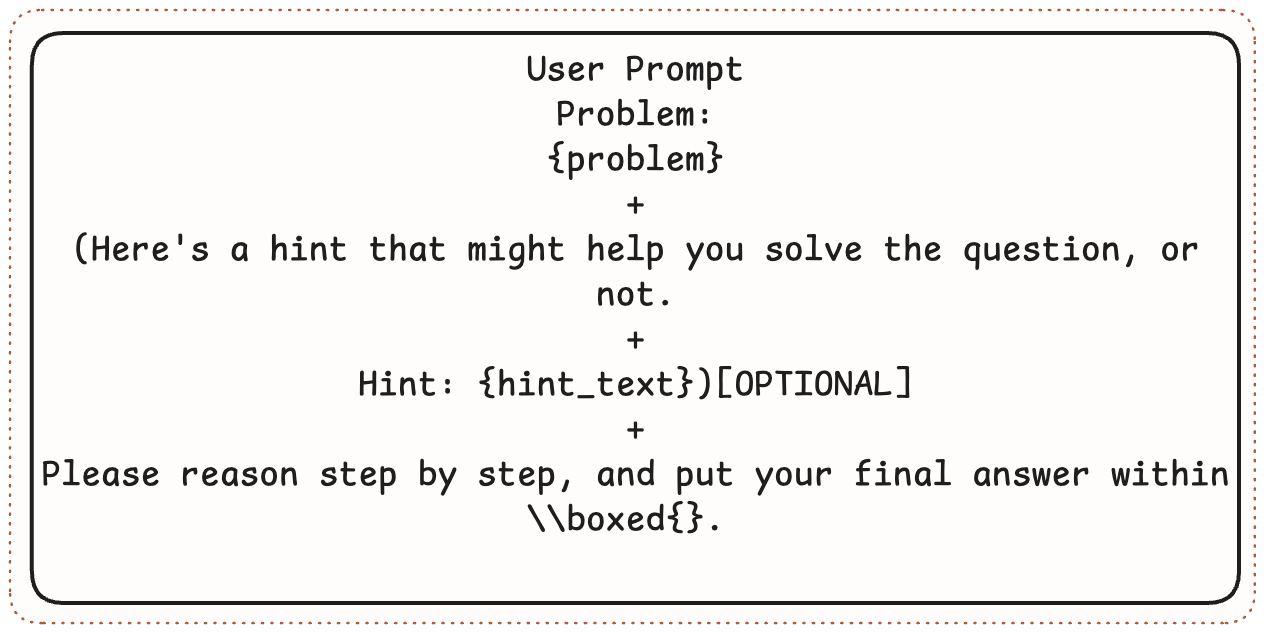}
    \caption{The Reasoner prompt}
    \label{fig:prompt_reasoner}
\end{figure}

\end{document}